\documentclass[11pt]{article}

\usepackage[preprint]{acl}

\usepackage{times}
\usepackage{latexsym}

\usepackage[T1]{fontenc}
\usepackage[utf8]{inputenc}

\usepackage{microtype}

\usepackage{inconsolata}

\usepackage{graphicx}

\usepackage{booktabs, multirow}
\usepackage{hyperref}
\usepackage{xurl}
\usepackage{xcolor,colortbl}
\usepackage{fontawesome}
\usepackage{subcaption}
\usepackage{amsmath}
\usepackage{comment}
\usepackage{tabularx}
\usepackage{array}
\usepackage{enumitem}

\definecolor{tartunlp_violet}{HTML}{7268D8}
\definecolor{tartunlp_violet_light}{HTML}{D3C6F7}

\definecolor{tartunlp_red}{HTML}{EF6650}
\definecolor{tartunlp_red_light}{HTML}{F7D4CB}

\definecolor{tartunlp_yellow}{HTML}{E0B12B}
\definecolor{tartunlp_yellow_light}{HTML}{FFDA99}
\definecolor{tartunlp_gray_extralight}{HTML}{F4F4F4}


\title{EstLLM: Enhancing Estonian Capabilities in Multilingual LLMs via Continued Pretraining and Post-Training}

\author{
\textbf{Aleksei Dorkin\textsuperscript{1}}\thanks{Equal contribution.},
\textbf{Taido Purason\textsuperscript{1}}\footnotemark[1],
\textbf{Emil Kalbaliyev\textsuperscript{1}},
\textbf{Hele-Andra Kuulmets\textsuperscript{1}},
\\
\textbf{Marii Ojastu\textsuperscript{1}},
\textbf{Mark Fišel\textsuperscript{1}},
\textbf{Tanel Alumäe\textsuperscript{2}},
\\
\textbf{Eleri Aedmaa\textsuperscript{3}},
\textbf{Krister Kruusmaa\textsuperscript{3,4}},
\textbf{Kairit Sirts\textsuperscript{1}}
\\
\\
\textsuperscript{1} Institute of Computer Science, University of Tartu, Tartu, Estonia
\\
\textsuperscript{2} Department of Software Science, Tallinn University of Technology, Tallinn, Estonia
\\
\textsuperscript{3} Institute of the Estonian Language, Tallinn, Estonia 
\\
\textsuperscript{4}School of Humanities, Tallinn University, Tallinn, Estonia
\\
   \textbf{Email:} \texttt{firstname.lastname@\{ut,taltech,eki,tlu\}.ee}
}

\begin{document}
\maketitle
\begin{abstract}
Large language models (LLMs) are predominantly trained on English-centric data, resulting in uneven performance for smaller languages. We study whether continued pretraining (CPT) can improve Estonian capabilities in multilingual LLMs while preserving English and general reasoning performance. Using Llama 3.1 8B and Apertus 8B as base models, we apply CPT with Estonian-enriched multilingual replay, followed by mostly English supervised fine-tuning, preference optimization, and chat vector merging.
Evaluation on Estonian benchmarks, targeted pairwise human evaluation, and an Estonian Chatbot Arena-style setup shows consistent improvements in Estonian language competence, reasoning, translation, and instruction-following.
Although Apertus exhibits stronger Estonian capabilities before adaptation, the more English-centric Llama model achieves substantially larger gains after adaptation.
While some English capabilities regress relative to the original instruction-tuned models, chat vector merging substantially restores English instruction-following and reasoning performance.
These findings suggest that CPT with balanced multilingual replay and lightweight post-training alignment can substantially improve single-language capabilities in multilingual LLMs.
\end{abstract}

\section{Introduction}
Large language models (LLMs) are trained on multilingual web-scale corpora \citep{grattafiori2024llama3herdmodels,gemmateam2025gemma3technicalreport}. However, the training data distributions across languages remain highly imbalanced \citep{penedo2025fineweb2}, with English dominating most pretraining mixtures \citep{penedo2024fineweb}. As a result, performance across languages remains uneven, particularly for smaller languages with more limited representation \citep{xuan2025mmlu,liu2025maxife,singh2025global}.

These limitations have motivated targeted efforts to improve language-specific capabilities in pretrained open models \citep{rodriguez-etal-2025-continued,etxaniz-etal-2024-latxa,masala-etal-2024-vorbesti,joshi-etal-2025-adapting}. A common approach is to adapt existing multilingual LLMs via continued pretraining (CPT) \citep{zosa2025continued,rodriguez-etal-2025-continued, fujii2024continual} or fine-tuning on additional target-language data \citep{kuulmets-etal-2024-teaching,masala-etal-2024-vorbesti} to strengthen language-specific competence while preserving general capabilities acquired during large-scale pretraining \citep{ibrahim2024simple}.

\begin{figure}[t]
  \includegraphics[width=\columnwidth]{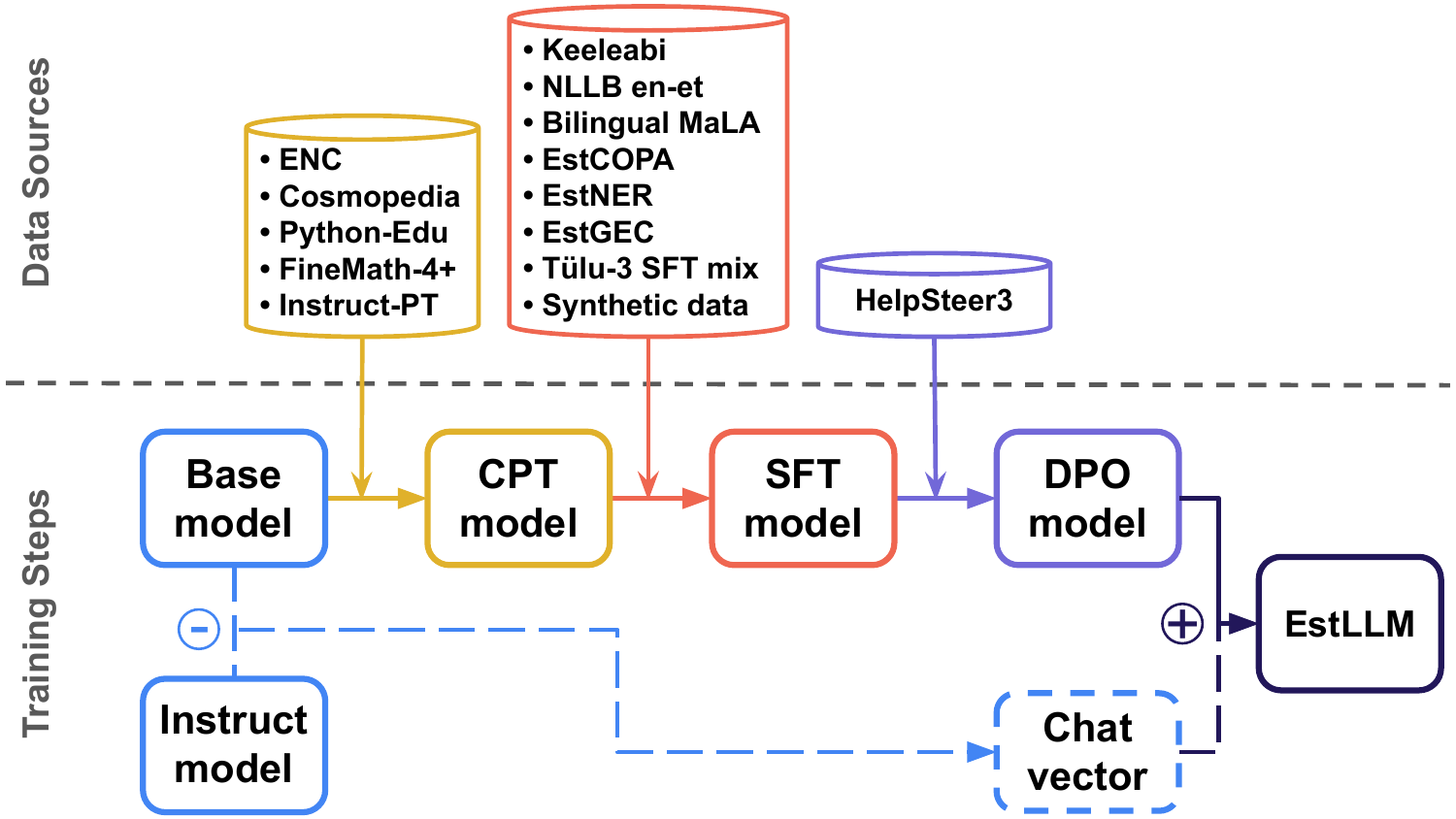}
  \caption{Overview of the EstLLM adaptation pipeline.}
  \label{fig:overview}
\end{figure}

In this paper, we investigate whether CPT can substantially improve Estonian capabilities in pretrained multilingual LLMs without compromising broader reasoning and instruction-following abilities.
We use Llama 3.1 8B \citep{grattafiori2024llama3herdmodels} and Apertus 8B \citep{hernandez2026apertus} (both non-instruction-tuned) as base models.
The models differ in their pretraining language distributions: Llama 3.1 was trained on predominantly English data, whereas Apertus allocated approximately 40\% of its pretraining data to non-English languages.
 We treat Estonian as a case study of language adaptation under multilingual data imbalance. 
Our focus is on empirically evaluating the effectiveness of established adaptation techniques and selected design choices, including base-model selection, multilingual vs monolingual CPT, data repetition, and model merging.

Our training pipeline (depicted in Figure~\ref{fig:overview}) consists of (i) continued pretraining on a data mixture that increases Estonian-language exposure while preserving general capabilities, followed by (ii) instruction tuning and preference optimization \citep{ouyang2022training}, primarily on English datasets, to restore instruction-following abilities. In addition, (iii) we apply chat vector merging \citep{huang-etal-2024-chat} to transfer instruction-following behavior from the instruction-tuned variants of the base model, i.e., Llama 3.1 8B Instruct and Apertus 8B Instruct, respectively.

The resulting EstLLM models are evaluated using (1) a comprehensive suite of Estonian benchmarks covering both understanding and generation tasks, (2) systematic pairwise human-evaluation, and (3) an Estonian-specific LMArena setup \citep{chiang2024chatbot} constructed for this study. Across automatic and comparative evaluation settings, both adapted models consistently outperform their original base models as well as their instruction-tuned counterparts on Estonian tasks. 
Moreover, the EstLLM model based on Llama 3.1 8B achieves the strongest overall performance  among models of comparable size, outperforming both  multilingual and English-centric alternatives on Estonian benchmarks \citep{martins2025eurollm9btechnicalreport,gonzalezagirre2025salamandratechnicalreport,liu2026ministral3,qwen2025qwen25technicalreport}. 
These findings show that CPT with a carefully constructed data mixture, combined with lightweight alignment techniques, can substantially improve single-language capabilities in pretrained multilingual LLMs.

Our contributions are: \textbf{(1) Adaptation pipeline:} a pipeline for adapting pretrained LLMs to Estonian via continued pretraining, post-training, and chat vector merging; \textbf{(2) Empirical insights:} evidence that English-centric models can outperform multilingual models after adaptation, and that chat vector merging improves both Estonian and English capabilities; \textbf{(3) Released models:} Estonian-focused models achieving top-tier arena-style human evaluation performance at the 8B scale;\footnote{\url{https://huggingface.co/collections/tartuNLP/estllm-llama-8b-models}, see Appendix~\ref{app:model-release}.} and \textbf{(4) Multi-turn human-evaluation benchmark:} we construct a new multi-turn prompt dataset enabling systematic human evaluation.

\section{Related work}
Enhancing language capabilities in LLMs involves multiple stages \citep{wang-etal-2025-language}, with the most prominent being continued pretraining (CPT) and post-training, while some works also explore vocabulary extension techniques \citep{dorkin-etal-2024-erzya, fujii2024continual, cui2023efficienteffectivetextencoding, wang-etal-2025-language, purason-etal-2026-teaching}.

CPT exposes pretrained models to target-language data to improve language competence before fine-tuning. Prior work explores language adaptation for Portuguese \citep{Pires_2023}, Estonian \citep{kuulmets-etal-2024-teaching}, Romanian \citep{masala-etal-2024-vorbesti}, Basque \citep{etxaniz-etal-2024-latxa}, Japanese \citep{fujii2024continual}, Lithuanian \citep{Nakvosas_2025}, and Hindi \citep{joshi-etal-2025-adapting}. Since updating all parameters may cause catastrophic forgetting \citep{jin-etal-2022-lifelong-pretraining}, several works use replay techniques \citep{shin2017continual,chaudhry2019tinyepisodicmemoriescontinual} that include portions of original training data during CPT \citep{scialom-etal-2022-fine,ibrahim2024simple}. 
Other studies incorporate related higher-resource languages for cross-lingual transfer \citep{purason-etal-2025-llms}, code and mathematical corpora \citep{aryabumi2025code,fujii2026rewriting,samuel-etal-2025-small, zosa2025continued}, or instruction datasets \citep{cheng2024instruction,rodriguez-etal-2025-continued}.
Machine-translated data has also been explored as a source for multilingual pretraining \citep{wang2025multilingual}.

Post-training aligns models with task objectives and user preferences, commonly through instruction tuning on translated or synthetically generated instruction datasets \citep{singh2024aya}. Prior work demonstrates effectiveness for Estonian \citep{kuulmets-etal-2024-teaching}, Chinese \citep{cui2023efficienteffectivetextencoding}, Basque \citep{sainz-etal-2025-instructing}, and low-resource Finno-Ugric languages \citep{purason-etal-2025-llms}. Several works further apply preference optimization to align model outputs with human preferences \citep{zosa2025continued,corral-etal-2025-pipeline} or model merging to recover instruction-following capabilities after CPT \citep{huang-etal-2024-chat,Kesgin_2024,antero-etal-2025-diploma}.

\begin{table}[t]
\centering
\small
\begin{tabular}{llr}
\toprule
\textbf{Domain} & \textbf{Source} & \textbf{Tokens} \\
\midrule
Estonian      & Estonian National Corpus       & 8.6B \\
English       & Cosmopedia v2           & 6.9B \\
Code          & Python-Edu                     & 3.3B \\
Math          & FineMath-4+                    & 9.5B \\
Instructions  & Instruct-PT           & 7.4B \\
\midrule
\textbf{Total} &                               & \textbf{35.7B} \\
\bottomrule
\end{tabular}
\caption{Continued pretraining data mixture. Token counts produced using the Llama 3.1 tokenizer.}
\label{tab:cpt-data-mix}
\end{table}

\section{Data}

\subsection{Continued Pretraining}

Strong distribution shifts during CPT tend to cause catastrophic forgetting, which can be mitigated through replay of original training domains \citep{ibrahim2024simple}. Since the original Llama 3.1 training data is not available, we approximate its composition using information from \citet{grattafiori2024llama3herdmodels}, who report that code and multilingual data constitute larger shares than in earlier Llama versions. This yields a mixture where Estonian occupies a modest share alongside English, code, and math---an approach consistent with \citet{zosa2025continued}, who adopt a similar balanced distribution when adapting Llama 3.1 to Finnish.

\paragraph{Estonian data.}
We use the Estonian National Corpus 2023 (ENC) \citep{koppel2021eestikeele,koppel2023eestikeele} as the primary Estonian data source. ENC was selected for its diversity of sources, including web data (e.g., the Estonian web, Wikipedia, and reference corpora) as well as literature and academic writing. However, the corpus is not consistently cleaned, filtered, or deduplicated. Therefore, we apply source-specific preprocessing steps described in Appendix~\ref{sec:ENC_preprocessing}.
After preprocessing, ENC contributes 8.6B tokens to the CPT mixture.
%and constitutes the largest component of the CPT mixture.

\paragraph{Continued Pretraining Mixture Design.}
\label{sec:cpt-data-mix}

To mitigate catastrophic forgetting, we construct a mixture combining Estonian data with English replay, code, math, and instruction-augmented corpora. We use Cosmopedia \citep{benallal2024smollmcorpus}, a high-quality synthetic English corpus, for English replay, Python-Edu \citep{benallal2024smollmcorpus} for code, FineMath-4+ \citep{allal2025smollm2smolgoesbig} for mathematics, and Instruct-PT \citep{cheng2024instruction}, which augments web documents with synthetic instructions. Following prior work, these components are included to support both replay and cross-domain transfer \citep{aryabumi2025code, cheng2024instruction, rodriguez-etal-2025-continued, fujii2026rewriting}.
Table~\ref{tab:cpt-data-mix} summarizes the resulting CPT mixture, totaling 35.7B tokens.

\paragraph{Warm-up phase.}
Before the main CPT stage, we perform a short warm-up phase to soften the distribution shift before exposure to the full Estonian-enriched mixture. The warm-up follows the same mixture design but places greater emphasis on curated high-quality data, including selected ENC subcorpora, English replay, code, math, instruction, and Estonian--English parallel data. Additional construction details are provided in Appendix~\ref{app:warmup}.

\subsection{Post-Training}
\label{sec:sft-data}

\paragraph{Supervised finetuning.} 
The supervised fine-tuning mixture was designed to both enable instruction-following in Estonian and preserve broader English capabilities.
Estonian data included Language help, a corpus of expert-authored Estonian language question-answer pairs, parallel translation data from NLLB \citep{nllbteam2022languageleftbehindscaling} and Bilingual MaLA \citep{ji2024emma500enhancingmassivelymultilingual}, and several Estonian NLP benchmarks \citep{kuulmets_estcopa_2022,sirts-2023-estonian} converted into instruction format. To increase the volume of Estonian instruction data, we additionally generated synthetic examples using Gemma 3 27B \citep{gemmateam2025gemma3technicalreport}, including translated prompts, Magpie-style synthetic prompt-response pairs \citep{xu2025magpie}, and translated documents. For English, we included Tülu-3 SFT mixture \citep{lambert2025tulu} and synthetic datasets from the EuroLLM project \citep{martins2025eurollm9btechnicalreport}. The total size of the SFT dataset was 766K instruction-response pairs. The instructions in the SFT dataset do not overlap with those used during CPT. Additional dataset details and statistics are provided in Appendix~\ref{app:sft-data}.

\paragraph{Preference Optimization.}
Because no Estonian-language preference dataset was available, we used the English HelpSteer3 preference dataset \citep{wang2026helpsteer3} for Direct Preference Optimization to address post-training alignment.
This follows the approach adopted for the Finnish Poro 2 model \citep{zosa2025continued}, where English-only HelpSteer3 performed comparably to its English-Finnish counterpart and outperformed other English preference datasets. 

\section{Training}
\subsection{Continued Pretraining} 

For CPT, sequences are packed to a length of 4096 tokens per example using best-fit bin packing \citep{garey1972worst} (see Appendix~\ref{app:cpt-preproc} for details). Unlike standard concatenate-then-split packing, this preserves document integrity by avoiding unnecessary truncation, which has been shown to improve downstream results \citep{ding2024fewer}. We use a trapezoidal learning rate schedule (warm-up--stable--decay), enabling flexible training durations. Our primary models are trained for a single epoch on the Estonian corpus, without data repetition. The total training token counts are 47.9B for Llama~3.1 and 48.8B for Apertus. Hyperparameters are reported in Appendix~\ref{app:cpt-hyperparams}.
We refer to the models produced in this stage as \texttt{Apertus-EstLLM-8B} and \texttt{Llama-3.1-EstLLM-8B}. 

For comparison, we also trained an Estonian-only variant of \texttt{Llama-3.1-8B}, excluding the English, code, mathematics, and instruction-augmented components of the multilingual mixture. 
We call it \texttt{Llama-3.1-EstLLM-8B-Mono}.
%The resulting model was subsequently post-trained and chat-vector merged using the same procedures as the multilingual variant.

\subsection{Post-Training}

\paragraph{Instruction-tuning} Instruction-following fine-tuning was performed via supervised fine-tuning (SFT) for one epoch. The original chat templates were retained without modification. The hyperparameters are listed in Appendix~\ref{app:posttrain-hyperparams}.

\paragraph{Preference Optimization}  Standard Direct Preference Optimization \citep{rafailov2023direct} with a frozen reference model was applied for one epoch using the HelpSteer3 dataset in its original English form. As the primary focus of this work is CPT, no systematic optimization of the preference learning stage was performed. We refer to the models produced in this stage as \texttt{Llama-EstLLM-8B-Instruct} and \texttt{Apertus-EstLLM-8B-Instruct}.

\paragraph{Chat vector merging} We adopt ChatVector merging \citep{huang-etal-2024-chat, antero-etal-2025-diploma} to transfer instruction-following behavior from the original instruction-tuned model. A chat vector $\Delta\theta$ is computed as the parameter difference between the instruction-tuned model $\theta_{\text{Instruct}}$ and its base model $\theta_{\text{Base}}$, and added to the continually pretrained model $\theta_{\text{CPT}}$:
%$$\Delta\theta = \theta_{\text{Instruct}} - \theta_{\text{Base}}$$
%$$\theta' = \theta_{\text{CPT}} + \alpha \Delta\theta$$
$$\Delta\theta = \theta_{\text{Instruct}} - \theta_{\text{Base}}, \qquad \theta' = \theta_{\text{CPT}} + \alpha \Delta\theta$$
We apply the chat vector to our post-trained model, following \citet{huang-etal-2024-chat}. The resulting model is referred to as \texttt{Llama-EstLLM-8B-Instruct\textsuperscript{$\spadesuit$}}.

\section{Evaluation}

The evaluation is structured to assess three central questions: (1) whether continued pretraining improves Estonian linguistic competence and knowledge, (2) whether Estonian instruction-following abilities improve after post-training, and (3) whether improvements in Estonian come at the cost of degraded capabilities in English or other domains.
We prioritize natively created datasets over machine-translated ones. Translated benchmarks have been shown to introduce translation noise and reduce cultural relevance, with localized benchmarks correlating better with human judgments than their translated counterparts \citep{lillepalu-alumae-2026-estonian, wu2025bitterlessonlearned2000}.

\paragraph{Estonian evaluation.} 
We evaluate linguistic competence using grammar correction, declension, and word meanings from the Estonian Native Large Language Model Benchmark \citep{lillepalu-alumae-2026-estonian}. Knowledge and reasoning are evaluated using trivia and national exams from the same benchmark, alongside GlobalPIQA \citep{chang2025globalpiqaevaluatingphysical}, Estonian WinoGrande \citep{ojastu-etal-2026-estonian}, XCOPA \citep{ponti-etal-2020-xcopa}, and Belebele \citep{bandarkar-etal-2024-belebele}. 
Instruction-following and robustness are evaluated using the Estonian translation of IFEval \citep{ifeval-et-doi} and a translated version of TruthfulQA \citep{thellmann2024multilingualllmevaluationeuropean}.
Machine translation quality is evaluated using FLORES-200 \citep{nllbteam2022languageleftbehindscaling} in Estonian-to-English and English-to-Estonian directions.
The grammar correction, declension, and word meanings datasets are reformulated as multiple-choice to enable more robust automatic evaluation.

\paragraph{Retention of general capabilities.}
To verify that models retain English capabilities as well as general knowledge and reasoning, we evaluate on the English versions of WinoGrande \citep{10.1145/3474381} and TruthfulQA \citep{lin-etal-2022-truthfulqa}, as well as PIQA \citep{Bisk_Zellers_LeBras_Gao_Choi_2020}, MMLU-Redux \citep{gema-etal-2025-done}, and GSM8K \citep{cobbe2021trainingverifierssolvemath}. English instruction-following abilities are evaluated with IFEval~ \citep{zhou2023instructionfollowingevaluationlargelanguage}.

\begin{table*}[!t]\centering

\small
\addtolength{\tabcolsep}{-3.5pt}
\begin{tabular}{lccccccccccc}\toprule
&\multicolumn{8}{c}{\textsc{ET Discriminative Benchmarks }(\textit{acc \%})} & &\multicolumn{2}{c}{\textbf{Average}} \\%\cmidrule{2-9}%\cmidrule{11-12}
\textbf{Model} &\textbf{Belebele} &\textbf{Exam} &\textbf{Grammar} &\textbf{Declension} &\textbf{Trivia} &\textbf{WinoGr.} &\textbf{XCOPA} &\textbf{GPIQA} &\textbf{} &\textbf{Score} &\textbf{Rank} \\\midrule
\multicolumn{12}{l}{\cellcolor[HTML]{f3f3f3}\textsc{Baselines:}} \\
EuroLLM-9B &69.9 &61.8 &66.3 &44.0 &37.1 &69.2 &71.2 &69.0 & &\cellcolor[HTML]{f3f3f3}61.1 &\cellcolor[HTML]{f3f3f3}5.1 \\
Ministral-3-8B-Base-2512 &26.3 &52.8 &64.1 &58.5 &31.6 &62.3 &56.0 &60.0 & &\cellcolor[HTML]{f3f3f3}51.4 &\cellcolor[HTML]{f3f3f3}7.4 \\
Llammas-base &38.7 &46.2 &53.8 &26.9 &33.6 &69.7 &68.6 &76.0 & &\cellcolor[HTML]{f3f3f3}51.7 &\cellcolor[HTML]{f3f3f3}6.9 \\
salamandra-7B &44.8 &50.5 &69.9 &26.8 &29.6 &67.3 &65.8 &71.0 & &\cellcolor[HTML]{f3f3f3}53.2 &\cellcolor[HTML]{f3f3f3}7.4 \\
Qwen2.5-7B &66.4 &45.5 &65.4 &45.2 &29.0 &53.0 &49.4 &54.0 & &\cellcolor[HTML]{f3f3f3}51.0 &\cellcolor[HTML]{f3f3f3}8.8 \\\midrule
Apertus-8B-2509 &76.8 &60.7 &78.9 &47.8 &32.9 &71.1 &67.8 &73.0 & &\cellcolor[HTML]{f3f3f3}63.6 &\cellcolor[HTML]{f3f3f3}4.5 \\
Llama-3.1-8B &67.0 &44.7 &65.8 &58.7 &30.0 &59.6 &53.2 &53.0 & &\cellcolor[HTML]{f3f3f3}54.0 &\cellcolor[HTML]{f3f3f3}7.5 \\\midrule
\multicolumn{12}{l}{\cellcolor[HTML]{f3f3f3}\textsc{Our models:}} \\
Apertus-EstLLM-8B &78.8 &63.6 &83.4 &52.3 &38.9 &75.2 &73.0 &79.0 & &\cellcolor[HTML]{f3f3f3}68.0 &\cellcolor[HTML]{f3f3f3}2.0 \\
Llama-3.1-EstLLM-8B &77.2 &57.0 &87.5 &61.9 &44.9 &74.0 &75.2 &78.0 & &\cellcolor[HTML]{f3f3f3}69.5 &\cellcolor[HTML]{f3f3f3}2.0 \\
Llama-3.1-EstLLM-8B-Mono &67.0 &54.6 &72.1 &48.2 &45.6 &72.8 &75.4 &77.0 & &\cellcolor[HTML]{f3f3f3}64.1 &\cellcolor[HTML]{f3f3f3}3.4 \\
\bottomrule
\end{tabular}
\caption{Base model performance comparison across Estonian language benchmarks.}\label{tab:base-models-et}

\end{table*}

\begin{table*}[!htp]\centering
\small
\addtolength{\tabcolsep}{-3pt}
\begin{tabular}{lccccccc@{\hspace{10pt}}cccc}\toprule
&\multicolumn{3}{c}{\textsc{EN Discriminative Bench.} (\textit{acc \%})} &\multicolumn{2}{c}{\textbf{Average}} & & &\multicolumn{2}{c}{\textsc{MT} \textit{(BLEU)}} &\textbf{Avg} \\%\cmidrule(lr){2-4}\cmidrule(lr){9-10}
\textbf{Model} &\textbf{Belebele} &\textbf{WinoGr.} &\textbf{MMLU-R.} &\textbf{Score} &\textbf{Rank} & & &\textbf{EN-ET} &\textbf{ET-EN} &\textbf{Rank} \\\midrule
\multicolumn{11}{l}{\cellcolor[HTML]{f3f3f3}\textsc{Baselines:}} \\
EuroLLM-9B &77.3 &73.2 &55.7 &\cellcolor[HTML]{f3f3f3}68.7 &\cellcolor[HTML]{f3f3f3}7.3 & & &29.0 &41.2 &\cellcolor[HTML]{f3f3f3}1.5 \\
Ministral-3-8B-Base-2512 &89.7 &77.1 &72.9 &\cellcolor[HTML]{f3f3f3}79.9 &\cellcolor[HTML]{f3f3f3}2.0 & & &12.6 &29.6 &\cellcolor[HTML]{f3f3f3}8.0 \\
Llammas-base &45.0 &72.0 &35.0 &\cellcolor[HTML]{f3f3f3}50.7 &\cellcolor[HTML]{f3f3f3}9.7 & & &22.0 &32.7 &\cellcolor[HTML]{f3f3f3}6.0 \\
salamandra-7b &53.1 &70.6 &44.9 &\cellcolor[HTML]{f3f3f3}56.2 &\cellcolor[HTML]{f3f3f3}9.3 & & &14.7 &18.2 &\cellcolor[HTML]{f3f3f3}8.0 \\
Qwen2.5-7B &91.2 &75.1 &75.0 &\cellcolor[HTML]{f3f3f3}80.4 &\cellcolor[HTML]{f3f3f3}2.7 & & &5.1 &27.5 &\cellcolor[HTML]{f3f3f3}9.0 \\\midrule
Apertus-8B-2509 &82.7 &76.1 &59.8 &\cellcolor[HTML]{f3f3f3}72.9 &\cellcolor[HTML]{f3f3f3}5.7 & & &25.0 &38.5 &\cellcolor[HTML]{f3f3f3}3.5 \\
Llama-3.1-8B &87.3 &78.5 &64.9 &\cellcolor[HTML]{f3f3f3}76.9 &\cellcolor[HTML]{f3f3f3}2.3 & & &13.5 &33.7 &\cellcolor[HTML]{f3f3f3}6.5 \\\midrule
\multicolumn{11}{l}{\cellcolor[HTML]{f3f3f3}\textsc{Our models:}} \\
Apertus-EstLLM-8B &84.3 &76.3 &62.5 &\cellcolor[HTML]{f3f3f3}74.4 &\cellcolor[HTML]{f3f3f3}4.7 & & &27.4 &37.4 &\cellcolor[HTML]{f3f3f3}3.5 \\
Llama-3.1-EstLLM-8B &87.0 &76.6 &62.7 &\cellcolor[HTML]{f3f3f3}75.4 &\cellcolor[HTML]{f3f3f3}3.7 & & &28.1 &36.8 &\cellcolor[HTML]{f3f3f3}3.5 \\
Llama-3.1-EstLLM-8B-Mono &78.6 &72.2 &50.2 &\cellcolor[HTML]{f3f3f3}67.0 &\cellcolor[HTML]{f3f3f3}7.7 & & &34.3 &18.1 &\cellcolor[HTML]{f3f3f3}5.5 \\
\bottomrule
\end{tabular}
\caption{Base model performance comparison on English language benchmarks and FLORES200 generative machine translation (MT) benchmark (BLEU).}
\label{tab:base-models-en-mt}
\end{table*}

\paragraph{Evaluation protocol.} Not all benchmarks are applicable to both base and instruction-tuned models. For example, IFEval is only meaningful for instruction-tuned models, while certain multiple-choice tasks are better suited for base model evaluation.
Base models are evaluated using log-likelihood scoring with LM Evaluation Harness \citep{lintang_sutawika_2024_12608602}, except for FLORES-200, which is evaluated generatively with deterministic decoding.
Base model evaluation uses benchmark-specific 0-, 3-, or 5-shot setting, detailed in Appendix~\ref{app:eval-details}.
Instruction-tuned models are evaluated generatively, since first-token probabilities can diverge substantially from the models' actual text outputs due to conversational preambles and refusals \citep{wang-etal-2024-answer-c}. Evaluation is conducted in a zero-shot setting with deterministic decoding.

\paragraph{Baseline models.} 
As baselines, we consider several 
%English-centric and multilingual 
models of similar size. In particular, in addition to the two base models used for adaptation, \texttt{Llama-3.1-8B} and \texttt{Apertus-8B}, we use \texttt{EuroLLM-9B} \citep{martins2025eurollm9btechnicalreport}, \texttt{Ministral-3-8B} \citep{liu2026ministral3}, \texttt{salamandra-7B} \citep{gonzalezagirre2025salamandratechnicalreport} and \texttt{Qwen2.5-7B} \citep{qwen2025qwen25technicalreport}. We also compare to \texttt{Llammas} \citep{kuulmets-etal-2024-teaching}---the \texttt{Llama-2-7B} adapted to Estonian.

\begin{table*}[ht]
\centering
\small
\setlength{\tabcolsep}{3pt}
\begin{tabularx}{\textwidth}{
    >{\raggedright\arraybackslash\hangindent=1em\hangafter=1}X*{11}{c}
}
\toprule
&&\multicolumn{5}{c}{\textsc{Knowledge and Reasoning}}&\multicolumn{3}{c}{\textsc{Language competence}}&\multicolumn{2}{c}{\textbf{Average}} \\
\cmidrule(lr){3-7} \cmidrule(lr){8-10} \cmidrule(lr){11-12}
\textbf{} &\textsc{IFEval}&\textbf{Wino-} &\textbf{} &\textbf{} &\textbf{Global} &\textbf{Truthful} & \textbf{} & \textbf{Declen-} & \textbf{Word} \\
\textbf{Model} &\textsc{ET}&\textbf{grande} &\textbf{Trivia} &\textbf{Exam} &\textbf{PIQA} &\textbf{QA} & \textbf{Grammar} & \textbf{sion} & \textbf{Meanings} &\textbf{Score} &\textbf{Rank}\\
\midrule
\multicolumn{12}{l}{\cellcolor[HTML]{f3f3f3}\textsc{Baselines:}} \\
EuroLLM-9B-Instruct &54.0&58.5 &37.4 &55.9 &55.0 &28.9 & 76.4 & 36.7 & 92.6 &\cellcolor[HTML]{f3f3f3}55.0 &\cellcolor[HTML]{f3f3f3}5.2\\
Ministral-3-8B-Instruct-2512 &48.9&58.1 &31.3 &50.1 &48.0 &35.3& 56.2 & 48.3 & 84.0 &\cellcolor[HTML]{f3f3f3}51.1 &\cellcolor[HTML]{f3f3f3}7.4 \\
Apertus-8B-Instruct-2509 &54.8&51.1 &34.5 &55.2 &51.0 &36.6 & 51.2 & 36.6 & 90.3 &\cellcolor[HTML]{f3f3f3}51.3 &\cellcolor[HTML]{f3f3f3}7.7\\
salamandra-7B-Instruct &51.9&28.8 &28.7 &35.6 &55.0 &30.1 & 59.4 & 26.7 & 80.8 &\cellcolor[HTML]{f3f3f3}44.1 &\cellcolor[HTML]{f3f3f3}9.6  \\
Llama-3.1-8B-Instruct &38.0&54.0 &28.9 &50.0 &54.0 &43.7 & 65.7 & 41.7 & 83.4 &\cellcolor[HTML]{f3f3f3}51.0 &\cellcolor[HTML]{f3f3f3}7.7\\
Qwen2.5-7B-Instruct &49.9&54.7 &29.4 &49.1 &57.0 &41.1 & 59.8 & 41.4 & 79.8 &\cellcolor[HTML]{f3f3f3}51.4 &\cellcolor[HTML]{f3f3f3}7.6 \\
Llammas &35.2&50.4 &28.4 &36.5 &1.0 &20.3  & 52.9 & 22.9 & 53.3 &\cellcolor[HTML]{f3f3f3}33.4 &\cellcolor[HTML]{f3f3f3}11.7\\\midrule
\multicolumn{12}{l}{\cellcolor[HTML]{f3f3f3}\textsc{Our models:}} \\
Llama-EstLLM-8B-Instruct &51.7&58.1 &42.5 &50.9 &63.0 &35.3 & 69.2 & 51.9 & 95.7 &\cellcolor[HTML]{f3f3f3}57.6 &\cellcolor[HTML]{f3f3f3}4.2\\
Llama-EstLLM-8B-Instruct\textsuperscript{$\spadesuit$} &61.4 &64.4 &42.9 &63.3 &68.0 &37.9 & 83.1 & 57.8 & 96.2 &\cellcolor[HTML]{f3f3f3}63.9 &\cellcolor[HTML]{f3f3f3}1.3\\
Llama-EstLLM-8B-Instruct-Mono\textsuperscript{$\spadesuit$} &48.1 &55.4 &38.4 &51.1 &28.0 &30.7 & 61.1 & 54.3 & 92.3 &\cellcolor[HTML]{f3f3f3}51.0 &\cellcolor[HTML]{f3f3f3}6.4\\
Apertus-EstLLM-8B-Instruct &46.7&54.7 &35.8 &56.5 &63.0 &37.0 & 64.6 & 42.1 & 91.8 &\cellcolor[HTML]{f3f3f3}54.7 &\cellcolor[HTML]{f3f3f3}5.7\\
Apertus-EstLLM-8B-Instruct\textsuperscript{$\spadesuit$}&56.1& 59.8 & 35.0 & 60.2 & 64.0 & 43.0 & 71.3 & 43.3 & 94.4 &\cellcolor[HTML]{f3f3f3}58.6 &\cellcolor[HTML]{f3f3f3}3.0\\
\bottomrule
\end{tabularx}
\caption{Estonian benchmark results of instruction-following models. \textsuperscript{$\spadesuit$} denotes chat-vector merging.}
\label{tab:est_autoeval_inst}
\end{table*}
\begin{table*}[ht]
\small
\centering
\setlength{\tabcolsep}{3pt}
\begin{tabularx}{\textwidth}{
    >{\raggedright\arraybackslash\hangindent=1em\hangafter=1}X*{8}{c}
}
\toprule
&\textsc{IFEval}&\multicolumn{5}{c}{\textsc{Knowledge and Reasoning}} &\multicolumn{2}{c}{\textbf{Average}}\\
\cmidrule(lr){3-7} \cmidrule(lr){8-9}
\textbf{Model}&\textsc{EN} & \textbf{WinoGr.} & \textbf{GPIQA} & \textbf{TruthfulQA} & \textbf{MMLU-R.} & \textbf{GSM8K} &\textbf{Score} &\textbf{Rank}\\
\midrule
\multicolumn{9}{l}{\cellcolor[HTML]{f3f3f3}\textsc{Baselines:}} \\
EuroLLM-9B-Instruct &70.0& 50.6 & 58.0 & 29.6 & 57.4 & 59.4 & \cellcolor[HTML]{f3f3f3}54.2 & \cellcolor[HTML]{f3f3f3}8.3 \\
Ministral-3-8B-Instruct-2512 &68.5& 65.0 & 77.0 & 51.9 & 74.2 & 39.3 & \cellcolor[HTML]{f3f3f3}62.7 & \cellcolor[HTML]{f3f3f3}4.5\\
Apertus-8B-Instruct-2509 &78.1& 51.3 & 60.0 & 39.1 & 61.1 & 56.8 & \cellcolor[HTML]{f3f3f3}57.7 & \cellcolor[HTML]{f3f3f3}6.3\\
Llama-3.1-8B-Instruct &81.1& 56.3 & 76.0 & 52.4 & 69.6 & 77.1 & \cellcolor[HTML]{f3f3f3}68.8 & \cellcolor[HTML]{f3f3f3}3.2\\
Llammas &43.7& 49.8 & 0.0 & 19.7 & 34.2 & 14.6 & \cellcolor[HTML]{f3f3f3}27.0 & \cellcolor[HTML]{f3f3f3}11.5\\
salamandra-7B-Instruct &32.9& 40.3 & 63.0 & 27.2 & 51.8 & 0.1 & \cellcolor[HTML]{f3f3f3}35.9 & \cellcolor[HTML]{f3f3f3}10.7\\
Qwen2.5-7B-Instruct &79.5& 66.3 & 83.0 & 58.8 & 75.6 & 78.6 & \cellcolor[HTML]{f3f3f3}73.6 & \cellcolor[HTML]{f3f3f3}1.3\\\midrule
\multicolumn{9}{l}{\cellcolor[HTML]{f3f3f3}\textsc{Our models:}} \\
Llama-EstLLM-8B-Instruct &75.3& 60.8 & 71.0 & 36.6 & 63.9 & 72.0 & \cellcolor[HTML]{f3f3f3}63.3 & \cellcolor[HTML]{f3f3f3}4.8\\
Llama-EstLLM-8B-Instruct\textsuperscript{$\spadesuit$} &81.7 & 61.2 & 76.0 & 36.4 & 66.1 & 77.3 & \cellcolor[HTML]{f3f3f3}66.5 & \cellcolor[HTML]{f3f3f3}3.5\\
Llama-EstLLM-8B-Instruct-Mono\textsuperscript{$\spadesuit$} &66.9 & 55.1 & 24.0 & 29.9 & 46.7 & 48.7 & \cellcolor[HTML]{f3f3f3}45.2 & \cellcolor[HTML]{f3f3f3}9.3\\
Apertus-EstLLM-8B-Instruct &66.4& 53.5 & 56.0 & 36.5 & 59.4 & 52.8 & \cellcolor[HTML]{f3f3f3}54.1 & \cellcolor[HTML]{f3f3f3}8.5\\
Apertus-EstLLM-8B-Instruct\textsuperscript{$\spadesuit$} &71.0& 57.0 & 69.0 & 41.7 & 59.5 & 55.9 & \cellcolor[HTML]{f3f3f3}59.0 & \cellcolor[HTML]{f3f3f3}5.8\\
\bottomrule
\end{tabularx}
\caption{English benchmark results for instruction-tuned models. \textsuperscript{$\spadesuit$} denotes chat-vector merging.}
\label{tab:eng_knowledge}
\end{table*}

\section{Results}

\subsection{Continued pre-training (CPT) results}

\paragraph{Existing base models} 
Before CPT, multilingual models such as \texttt{EuroLLM-9B} and \texttt{Apertus-8B} perform better on Estonian tasks (see Table~\ref{tab:base-models-et}). 
In contrast, \texttt{Qwen2.5-7B}, \texttt{Llama-3.1-8B}, and \texttt{Ministral-3-8B-Base} achieve stronger English performance (see Table~\ref{tab:base-models-en-mt}), but perform worse on Estonian discriminative tasks and machine translation compared to \texttt{EuroLLM-9B} and \texttt{Apertus-8B}.

\paragraph{Preliminary experiments.} Preliminary CPT experiments indicated that ENC outperformed FineWeb-2 as the Estonian corpus. Repeating the Estonian CPT data for multiple epochs yielded only marginal gains in automatic evaluation on Estonian benchmarks despite substantially higher training cost. We therefore use ENC and a single training epoch in subsequent experiments.
Those choices are explored in depth in Appendix~\ref{sec:appenidx-prelim-cpt}.

\paragraph{Continued pretraining.}   
Before CPT, the multilingual Apertus performs better than the more English-centric Llama 3.1 on Estonian tasks (Table~\ref{tab:base-models-et}). 
After CPT, however, Llama 3.1 reaches comparable or stronger Estonian performance, suggesting that pre-CPT target-language capability might not best predict the target-language performance after CPT, consistent with the findings of \citet{yu-etal-2026-afriquellm}.
Mean relative gains across the eight Estonian discriminative benchmarks are larger for  Llama (+30.4\%) than for Apertus (+7.8\%). Although CPT slightly reduces English performance for the Llama-based model (see Table~\ref{tab:base-models-en-mt}), it still achieves stronger English results overall than the Apertus-based model.
Compared to other models of similar size, our models achieve the highest discriminative performance on Estonian. 
On English discriminative tasks, our models are only outperformed by \texttt{Qwen2.5-7B}, \texttt{Ministral-3-8B-Base}, and \texttt{Llama-3.1-8B}.

\subsection{Post-training results}

\paragraph{Instruction-tuning (SFT+DPO)} 
Instruction-tuned models were evaluated generatively on instruction-following, Estonian language competence, and knowledge and commonsense reasoning. 
The Llama-based EstLLM model demonstrates consistent improvements over the original \texttt{Llama-3.1-8B-Instruct} in Estonian across all evaluation categories, particularly in language competence and reasoning tasks, while the Apertus-based model shows more modest gains relative to \texttt{Apertus-8B-Instruct} (see Table~\ref{tab:est_autoeval_inst}). Although Apertus performed better on Estonian tasks before adaptation, the Llama-based model achieves stronger overall Estonian performance after CPT and post-training.
These gains, however, come with trade-offs in English for both model families. Compared to their respective original instruction-tuned models, both EstLLM variants exhibit lower performance on five of the six English benchmarks, with WinoGrande being the only exception (see Table~\ref{tab:eng_knowledge}).
This pattern---improvements in Estonian accompanied by declines on several English benchmarks---points to a tension between language-specific adaptation and the preservation of general-purpose capabilities that warrants further investigation.

\paragraph{Chat-vector merging.}  
We further evaluate chat vector merging for the Llama-based model.
%, yielding \texttt{Llama-3.1-EstLLM-8B-Instruct-CV}.
We exclude the embedding layer from the merge and find $\alpha=0.5$ to give the best results.
The merged model largely resolves the trade-offs observed in \texttt{Llama-EstLLM-8B-Instruct}. 
Instruction-following, measured with IFEval in both Estonian and English, improves substantially in both languages (Tables~\ref{tab:est_autoeval_inst} and~\ref{tab:eng_knowledge}), while Estonian language competence and reasoning further improve over both the base model \texttt{Llama-3.1-8B-Instruct} and the non-merged checkpoint \texttt{Llama-EstLLM-8B-Instruct} (Table~\ref{tab:est_autoeval_inst}). 
Crucially, regressions on English knowledge and reasoning benchmarks present in \texttt{Llama-EstLLM-8B-Instruct} are substantially reduced, with the merged model matching or exceeding the base \texttt{Llama-3.1-8B-Instruct} on most English tasks (Table~\ref{tab:eng_knowledge}).  
The Estonian-only CPT variant achieves lower average scores than its multilingual CPT counterpart on both Estonian and English benchmarks, consistent with the results observed for base models (the models with Mono-suffix in Tables~\ref{tab:base-models-et}--\ref{tab:eng_knowledge}).
Together, these results suggest that chat vector merging can restore general-purpose capabilities affected during language adaptation, while preserving and even compounding the Estonian-specific gains.

\subsection{Targeted Pairwise Human Evaluation}

Because automatic evaluation provides limited coverage of generative capabilities, we additionally conducted targeted pairwise human evaluation of post-trained models on a new curated multi-turn Estonian benchmark derived from translated LMSYS-Chat \citep{zheng2024lmsys} and SMUGRI-MT-Bench \citep{purason-etal-2025-llms}. The new benchmark contains 80 two-turn conversations spanning planning and scheduling, specific-format writing, mathematics, and logical reasoning. Seventy-four Estonian-speaking annotators compared model pairs on fluency, grammar, sequential instruction-following, correctness/usefulness, and overall preference. Evaluations were based on the complete two-turn conversation. 
Model pairs were selected such that each comparison isolates a single design choice, enabling targeted evaluation of specific adaptation components and training decisions. 
Full dataset construction and evaluation process details are provided in Appendix~\ref{app:human_eval}.

\begin{figure}[h!]
    %%\centering
    \captionsetup[subfigure]{
        justification=raggedright,
        singlelinecheck=false
    }

    \begin{subfigure}[t]{\columnwidth}
        %%\centering
        \includegraphics[width=\linewidth]{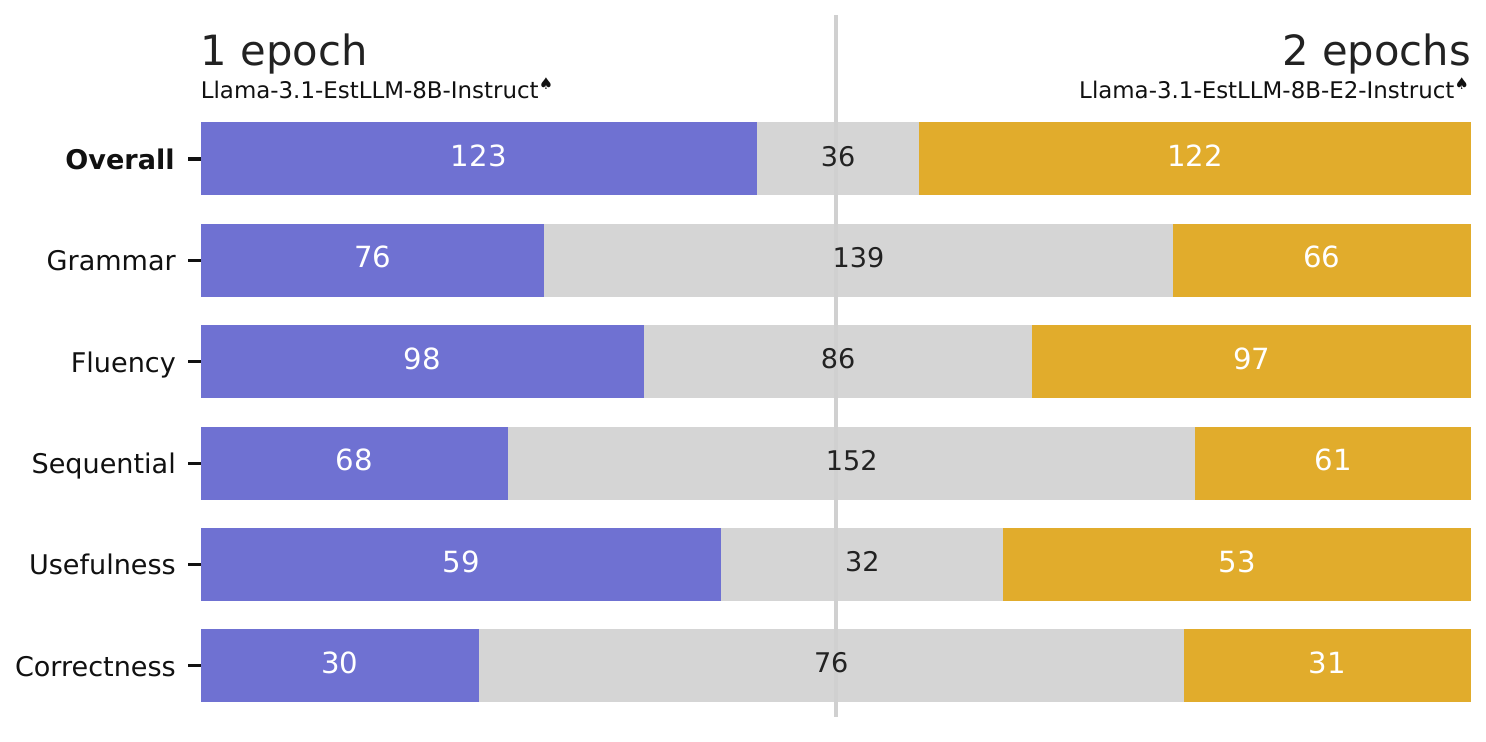}
        \caption{\textit{Does repeating Estonian data improve results?}}
        \label{fig:human-eval-epochs}
    \end{subfigure}
    
    \vspace{0.5em}

    \begin{subfigure}[t]{\columnwidth}
        %\centering
        \includegraphics[width=\linewidth]{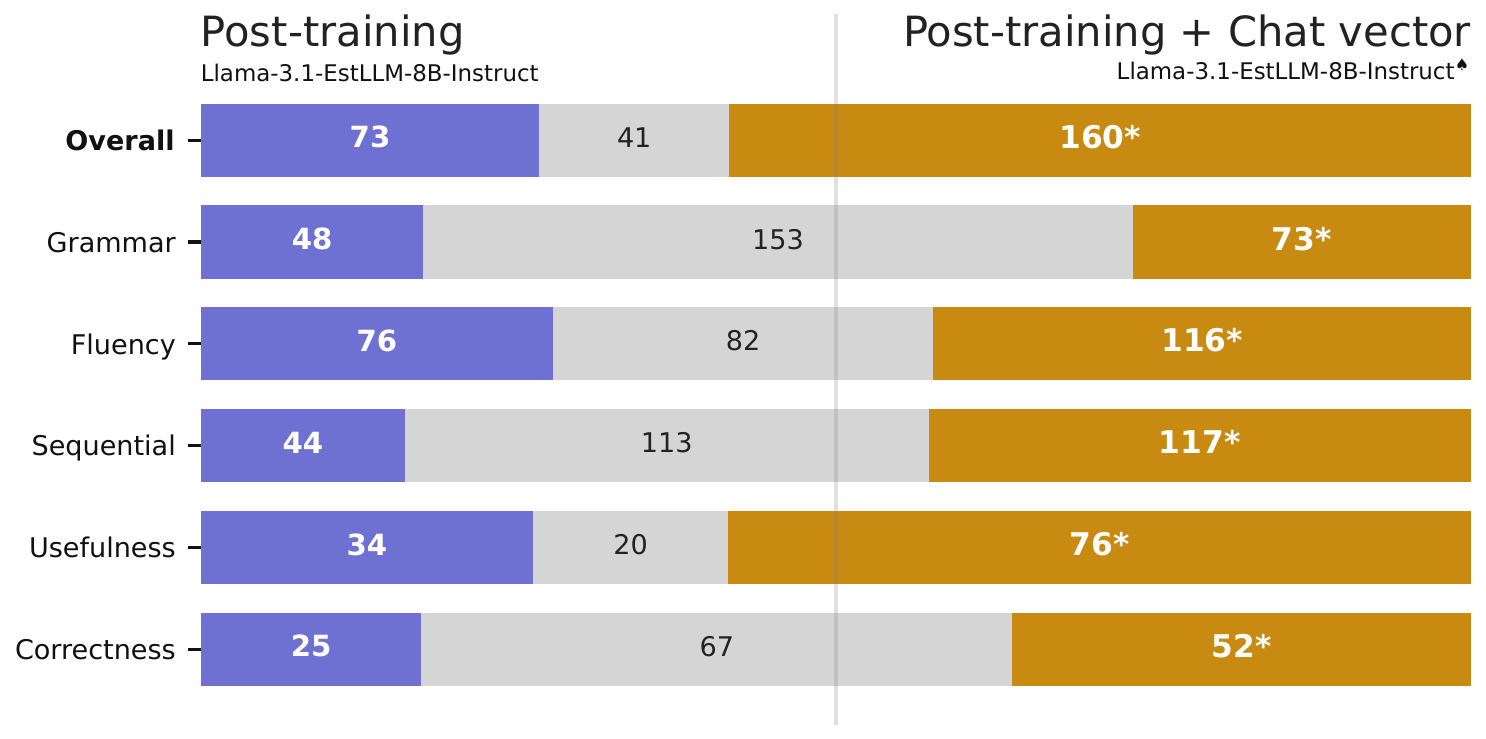}
        \caption{\textit{Does chat vector improve results?}}
        \label{fig:human-eval-cv}
    \end{subfigure}

    \vspace{0.5em}

    \begin{subfigure}[t]{\columnwidth}
        %\centering
        \includegraphics[width=\columnwidth]{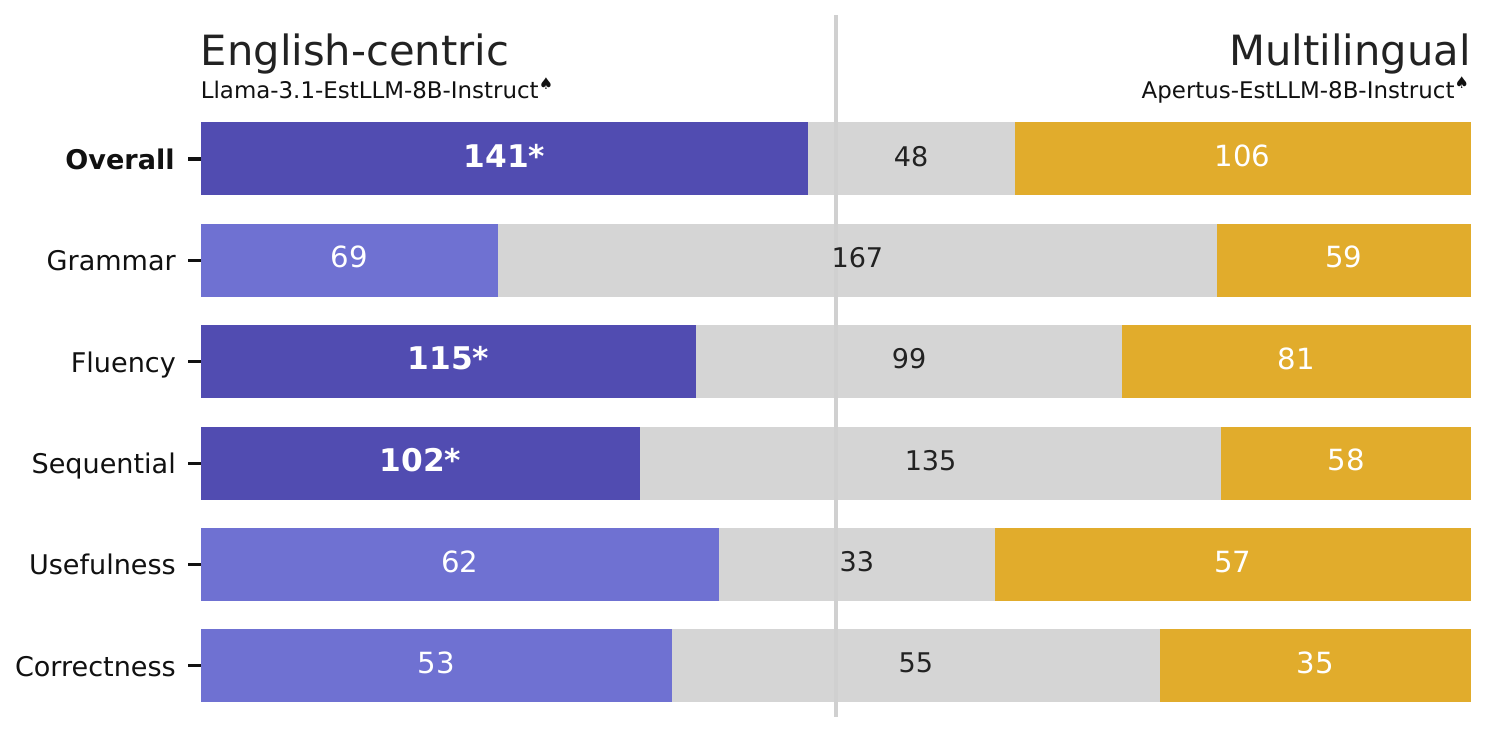}
        \caption{\textit{Does a massively multilingual model perform better than an English-centric model after CPT?}}
        \label{fig:human-eval-base-model}
    \end{subfigure}

    \vspace{0.5em}

    \begin{subfigure}[t]{\columnwidth}
        %\centering
        \includegraphics[width=\columnwidth]{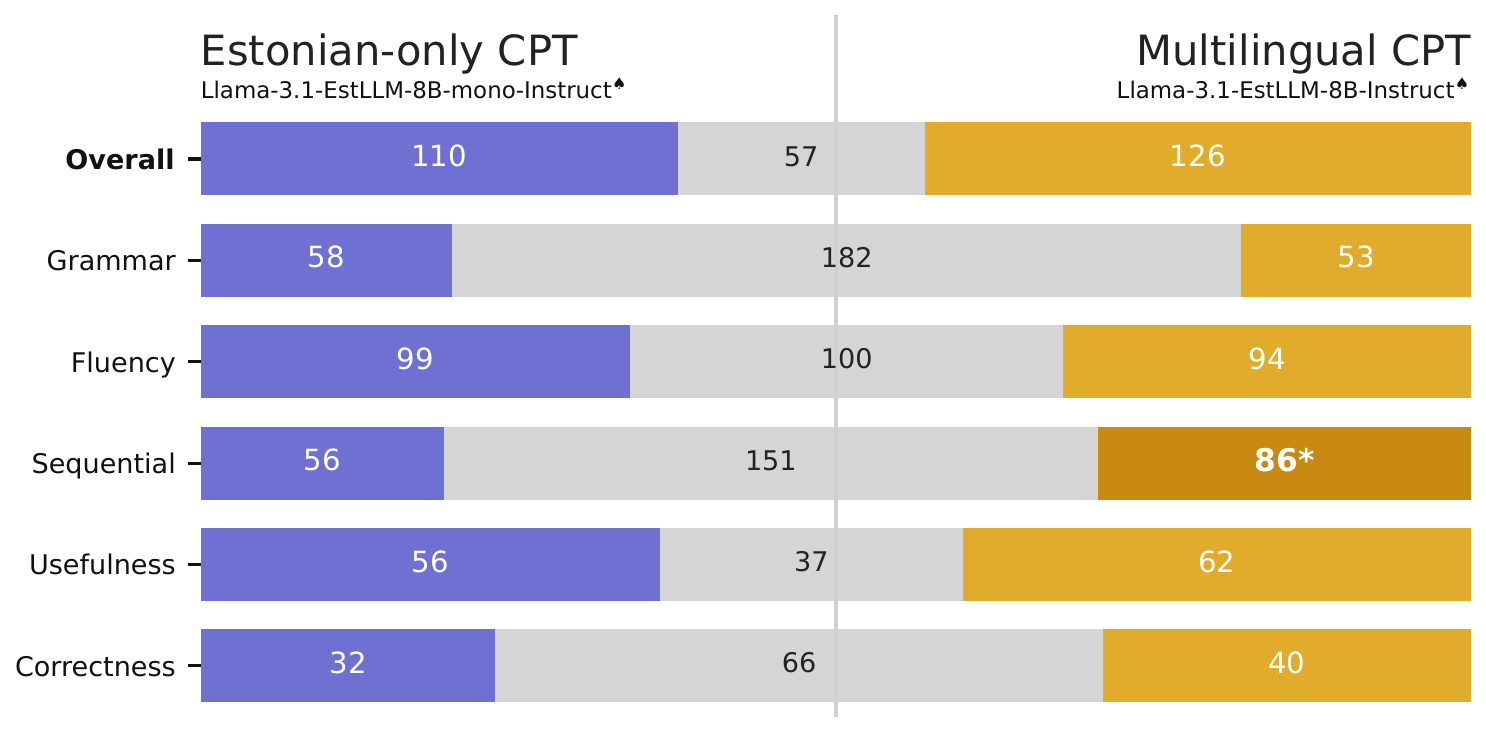}
        \caption{\textit{Does replaying English and code along Estonian data improve the results compared to Estonian-only CPT?}}
        \label{fig:human-eval-cv-replay}
    \end{subfigure}

    \caption{Pairwise human evaluation results (counts of \textit{left/tie/right} preference). Statistically significant preferences determined using bootstrap resampling are indicated with * ($\alpha=0.05$). \textsuperscript{$\spadesuit$} denotes chat-vector merging.}
    \label{fig:human-eval-full}
\end{figure}

\paragraph{\faQuestionCircle{} Does repeating Estonian data improve results?} Human evaluation shows no evidence that training for two epochs on the Estonian CPT data improves performance on any evaluation dimension relative to a single epoch (see Figure~\ref{fig:human-eval-epochs}).

\paragraph{\faQuestionCircle{} Does chat vector improve results?} 
Human evaluation shows that adding a chat vector significantly improves results in all categories (see Figure~\ref{fig:human-eval-cv}). This improvement was also apparent in automatic evaluation.

\paragraph{\faQuestionCircle{} Does a massively multilingual model perform better than an English-centric model after CPT?} 
%\paragraph{\faQuestionCircle{} Does a more multilingual base model benefit more from CPT than an English-centric model?}  
Human evaluation shows that adaptation starting from the more English-centric Llama 3.1 yields better results than adaptation starting from the massively multilingual Apertus (see Figure~\ref{fig:human-eval-base-model}). In addition to being preferred overall, the Llama-based model is also significantly preferred for fluency and sequential instruction-following.

\paragraph{\faQuestionCircle{} Does replaying English and code along Estonian data improve the results compared to Estonian-only CPT?} 
Human evaluation does not show a significant overall preference for multilingual CPT with replay over Estonian-only CPT. 
Among individual evaluation dimensions, multilingual CPT is only significantly better at sequential instruction following, while no significant differences are observed in other categories.

\subsection{Arena-Style Pairwise Human Evaluation}
\label{sec:arena-eval}

In addition to benchmark-based evaluation, we adopt a more holistic Chatbot Arena-style evaluation \citep{chiang2024chatbot} based on pairwise human preferences using a public web interface providing access to multiple proprietary and open-source Estonian-capable models, including instruction-tuned variants of EstLLM models. 
Users could submit arbitrary prompts, without predefined tasks or templates. For each prompt, anonymized responses from two models were presented side by side, and users selected the preferred answer.  
A ranked leaderboard was constructed from pairwise votes using a Bradley-Terry model, with ranks determined using bootstrapped confidence intervals (\citealp{chiang2024chatbot}; see Appendix~\ref{app:arena-results}).

A snapshot of the top-5-rank open models is in Table~\ref{tab:arena_eval_small}.
The instruction-tuned EstLLM variants rank competitively, with the chat-vector-merged Llama-based model sharing first rank alongside several substantially larger models, while the non-merged variant ranks within the top group.
Other open models of comparable size rank below our 8B model, and the directly comparable \texttt{Llama-3.1-8B-Instruct} does not appear among this top group. For more details see Appendix~\ref{app:arena-results}.

\begin{comment}
\begin{table}[!htp]\centering
%\resizebox{\textwidth}{!}{ % use this if the table is too large
\small
\addtolength{\tabcolsep}{-3pt}
\begin{tabular}{clrc}
\toprule
\bf Rank & \bf Model & \bf Size & \bf Score \tiny{95\% CI}\\
\midrule
 1 & DeepSeek-V3 (0324) & 685B & 1457 \tiny{$-$28/$+$41} \\
 1 & Kimi-K2-Instruct & 1T & 1442 \tiny{$-$40/$+$36}  \\
 1 & Llama 4 Maverick & 402B & 1432 \tiny{$-$31/$+$38} \\
 1 & Gemma-3-27B-it & 27B & 1413 \tiny{$-$38/$+$35}   \\
 \rowcolor{lightgray}1 & Llama-EstLLM-8B-Instruct-CV & 8B & 1384 \tiny{$-$47/$+$53}  \\
  \rowcolor{lightgray}2 & Llama-EstLLM-8B-Instruct & 8B & 1376 \tiny{$-$40/$+$39 } \\
 2 & Llama 4 Scout & 109B & 1374 \tiny{$-$37/$+$32}  \\
 4 & Meta-Llama-3.1-405B-Instruct & 405B & 1358 \tiny{$-$35/$+$38}  \\
 4 & Qwen3-235B-A22B & 235B & 1347 \tiny{$-$40/$+$36}  \\
  \rowcolor{lightgray}4 & Apertus-EstLLM-8B-Instruct & 8B & 1321 \tiny{$-$45/$+$57}  \\
 4 & Gemma-3-12B-it & 405B & 1317 \tiny{$-$65/$+$59}  \\
 5 & Llama-3.3-70B-Instruct & 70B & 1329 \tiny{$-$33/$+$37}  \\
 5 & EuroLLM-9B-Instruct & 9B & 1328 \tiny{$-$32/$+$43}  \\
 8 & Apertus-8B-Instruct-2509 & 8B & 1285 \tiny{$-$46/$+$42}  \\
 8 & Meta-Llama-3.1-70B-Instruct & 70B & 1281 \tiny{$-$46/$+$52}  \\
\bottomrule
\end{tabular}
\caption{Snapshot of the open models leaderboard from the AI Barometer---a Chatbot Arena style evaluation page in Estonian---as of 19.02.2026.}
\label{tab:arena_eval}
\end{table}
\end{comment}

\begin{table}[t]\centering
\small
\addtolength{\tabcolsep}{-3pt}
\begin{tabular}{cllc}
\toprule
\bf Rank & \bf Model & \bf Size & \bf Score \\
\midrule
 1 & DeepSeek-V3 (0324) & 685B & 1457  \\
 1 & Kimi-K2-Instruct & 1T & 1442  \\
 1 & Llama 4 Maverick & 402B & 1432  \\
 1 & Gemma-3-27B-it & 27B & 1413   \\
    \rowcolor{tartunlp_gray_extralight}1 & Llama-EstLLM-8B-Instruct\textsuperscript{$\spadesuit$} & 8B & 1384  \\
\rowcolor{tartunlp_gray_extralight}2 & Llama-EstLLM-8B-Instruct & 8B & 1376  \\
 2 & Llama 4 Scout & 109B & 1374  \\
 4 & Meta-Llama-3.1-405B-Instruct & 405B & 1358  \\
 4 & Qwen3-235B-A22B & 235B & 1347  \\
 \rowcolor{tartunlp_gray_extralight}4 & Apertus-EstLLM-8B-Instruct & 8B & 1321   \\
 4 & Gemma-3-12B-it & 12B & 1317  \\
 5 & Llama-3.3-70B-Instruct & 70B & 1329  \\
 5 & EuroLLM-9B-Instruct & 9B & 1328   \\
\bottomrule
\end{tabular}
\caption{Snapshot of the open-model leaderboard from an Estonian Chatbot Arena style evaluation platform. \textsuperscript{$\spadesuit$} denotes chat-vector merging.}
\label{tab:arena_eval_small}
\end{table}

\section{Discussion}

\paragraph{Data repetition in CPT.} 
We observed no substantial improvement beyond a single epoch over approximately 35B tokens of Estonian-enriched data. This contrasts with recent language-specific adaptation efforts, where CPT data is often repeated \citep{samuel-etal-2025-small, etxaniz-etal-2024-latxa, corral-etal-2025-pipeline}, apparently inspired by findings of \citet{JMLR:v26:24-1000}, who found that data repetition improves performance in data-constrained pretraining from scratch.
However, CPT differs fundamentally from training from scratch as it begins with models that already possess semantic and syntactic capabilities.  
Existing adaptation studies also rarely isolate the effect of repetition itself, adopting multi-epoch training without explicit validation of downstream gains. Our results suggest that, at least at the 8B scale and for the data volumes considered here, a single epoch may suffice, with further gains more likely to come from improving data quality or source diversity than additional passes over the same corpus.

\paragraph{Multilingual vs. Estonian-only CPT.}
When comparing the effect of monolingual vs multilingual CPT mixture, the model trained with Estonian-only CPT data performs worse on Estonian benchmarks both in base model evaluation as well as after post-training, although human evaluation did not detect a significant overall preference between models trained with monolingual vs multilingual mixture. Also, the English benchmark results dropped considerably for the model trained with monolingual Estonian data, suggesting that monolingual training leads to catastrophic forgetting of English and perhaps also general reasoning capabilities. This may also account for the lower performance on Estonian benchmarks. In summary, our results show that multilingual CPT offers advantages over monolingual CPT even when the goal is to improve performance in a single target language.

\paragraph{Effectiveness of chat vector merging.}
One of the more surprising findings of our study is the effectiveness of chat vector merging \citep{huang-etal-2024-chat}. As expected, the merged model retains stronger English instruction-following capabilities than the non-merged post-trained variant by incorporating behavior from Llama 3.1 8B Instruct. However, it also improves Estonian performance, despite the original Llama 3.1 8B Instruct itself performing poorly on Estonian tasks. The mechanism behind this effect remains unclear, but one possible explanation is that the chat vector captures language-independent instruction-following and reasoning patterns that complement the Estonian-specific knowledge acquired during CPT. Chat vector merging offers a computationally inexpensive way to improve both source- and target-language performance after language-specific adaptation.

\paragraph{Base model sensitivity to fine-tuning.}
Our results suggest that adaptation strategies are not equally effective across base models. Despite stronger multilingual and Estonian capabilities pre-adaptation, Apertus derives smaller gains from CPT and post-training than the more English-centric Llama 3.1 8B. This indicates that adaptation effectiveness depends not only on target-language competence before adaptation, but also on the original pretraining distribution and model specialization.

\section{Conclusions}
This work investigated whether CPT can substantially improve Estonian capabilities in pretrained multilingual LLMs while preserving English general reasoning performance. 
CPT on a multilingual training mixture with increased Estonian exposure yielded consistent improvements across Estonian benchmarks, targeted pairwise human evaluation, and an Estonian Chatbot Arena-style setup.
Although the multilingual Apertus baseline initially outperformed Llama 3.1 on Estonian tasks, the more English-centric Llama model exhibited substantially larger gains after adaptation, suggesting that initial target-language performance is not necessarily a reliable predictor of adaptation potential.
Our experiments further indicated the limited benefit of repeating target-language data for additional epochs at this scale.
The comparison with Estonian-only CPT further shows that a multilingual training mixture achieves stronger Estonian performance while better preserving English capabilities.
Our experiments also demonstrated the effectiveness of chat vector merging for restoring instruction-following behavior after language adaptation. 
The findings provide practical guidance for language-specific adaptation of multilingual LLMs and suggest that similar strategies may generalize across base models, languages, and model scales.

\section*{Limitations}
This study has several limitations. First, the data mixtures used for continued pretraining and post-training were not systematically optimized, and we did not conduct full ablations over mixture composition due to computational constraints. Second, all experiments were conducted at the 8B parameter scale, and it remains to be verified whether the same adaptation dynamics hold for substantially larger models.

Third, while we observe consistent improvements across Estonian benchmarks relative to the base model, our experiments do not conclusively identify which combination of training datasets is sufficient or optimal. 
Our comparisons examined the choice of Estonian-language data sources, data repetition, and monolingual versus multilingual CPT, but did not isolate the contributions of individual components of the multilingual mixture.
The benchmark improvements are clearly reflected on the leaderboard, but the relationship between specific data sources and individual task performance remains difficult to disentangle. A more systematic exploration of dataset composition---including the role of domain-specific, synthetic, and parallel data---is an important direction for future work.

Fourth, although the evaluation includes both controlled pairwise human evaluation and an open Arena-style evaluation, we did not systematically investigate longer conversational interaction patterns. The Arena-style evaluation relied on organically collected user interactions and freely submitted prompts, without following a controlled experimental protocol.

Next, we did not specifically evaluate Estonian cultural knowledge, as no established benchmark targeting this capability was available. Although some existing benchmark items and Arena prompts may contain culturally specific content, systematically identifying and evaluating this content would require substantial manual analysis, which was beyond the scope and resources of this study.

Although chat vector merging may transfer some safety-related behaviors from the original instruction-tuned models, we did not perform dedicated safety alignment or safety-focused evaluation in Estonian. Consequently, the resulting models should not be assumed to inherit the safety properties of the original instruction-tuned systems, particularly in lower-resource language settings where such behaviors may transfer inconsistently.

Finally, although we compare Estonian-only training with the full mixed-data setting, these experiments do not disentangle the effects of mixture composition from total training-token budget. Because the Estonian-only condition already used all available Estonian tokens, matching the token count of the mixed-data setting would have required either repeating Estonian data or downsampling the mixed-data run, both of which would introduce additional confounds. Consequently, the individual contributions of English replay, code, mathematics, and instruction-augmented data to preserving broader capabilities remain only partially understood.

\section*{Acknowledgments}
This work was supported by the National Programme for Estonian Language Technology (project EKTB104) and the Estonian Centre of Excellence in Artificial Intelligence (EXAI; grant TK213), both funded by the Estonian Ministry of Education and Research, and partially supported by the Estonian Research Council (grant PSG721).
We acknowledge the Estonian Scientific Computing Infrastructure (ETAIS, Estonia) for awarding this project access to the LUMI supercomputer, owned by the EuroHPC Joint Undertaking and hosted by CSC (Finland) and the LUMI consortium, through Estonia's national LUMI resource allocation.
We thank the students enrolled in the Spring 2026 bachelor's-level Artificial Intelligence course at the Institute of Computer Science, University of Tartu, for their participation as annotators in the human evaluation.

\bibliography{custom}

\appendix

\section{Data}
\subsection{Estonian data preprocessing}
\label{sec:ENC_preprocessing}

We use the DataTrove library \citep{penedo2024datatrove} for quality filtering, language filtering, and deduplication of the Estonian National Corpus (ENC).

\textbf{Normalization.} ENC is normalized using Unicode NFKC normalization to ensure consistent representation of characters across the corpus.

\textbf{Cleaning.} ENC is cleaned using a regex-based preprocessing pipeline designed to address source-specific noise. HTML, XML, and MediaWiki template tags, bracketed annotations, and user discussion metadata are removed from web and academic subcorpora. Where necessary, paragraph structure was preserved by converting selected tags into newline characters. Literature, Estonian Reference, and Balanced subcorpora are left unchanged.

\textbf{Quality Filtering.} After cleaning, web-based data are filtered based on the heuristics adapted from the Gopher \citep{rae2021scalinglanguagemodelsmethods} and C4 \citep{JMLR:v21:20-074} quality filtering pipelines. These filters target documents that are likely to contain low-quality or non-linguistic content. Documents shorter than four words or containing curly-bracket markup are removed. Additional constraints exclude texts with abnormal lexical statistics, including average word length outside the range of 3--12 characters, high symbol-to-word ratios (> 0.1), excessive non-alphabetic tokens (> 70\%), or large proportions of bullet-point lines (> 90\%) or ellipsis lines (> 30\%).

\textbf{Language Filtering.} All corpus texts are passed to language filtering using a FastText-based language identification model \citep{joulin2016fasttextzipcompressingtextclassification, joulin2017bag}. Documents predicted as Estonian with a score of at least 0.5 are retained.

\textbf{Deduplication.} Corpus documents are deduplicated using a 64-bit MinHash algorithm based on SHA-1. We use 5-grams with 14 buckets and 32 hashes per bucket to determine near-duplicate documents. All documents except a single sample are removed per duplicate.

\begin{table}[t]
\small
\centering
\begin{tabular}{llrr}
\toprule
\textbf{Domain} & \textbf{Source} & \textbf{Tokens} & \textbf{Share} \\
\midrule
Estonian & Curated ENC & 0.61B & 5.1\% \\
English & Cosmopedia & 6.99B & 58.0\% \\
Code & Python-Edu & 3.34B & 27.7\% \\
Math & NuminaMath-CoT & 0.43B & 3.6\% \\
Instructions & Magpie-Ultra & 0.30B & 2.5\% \\
Translation & NLLB (EN--ET) & 0.38B & 3.1\% \\
\midrule
\textbf{Total} & & \textbf{12.05B} & \textbf{100.0\%} \\
\bottomrule
\end{tabular}
\caption{Warm-up phase data mixture.}
\label{tab:warmup-mixture}
\end{table}

\begin{table*}[t]
\centering
\small
\begin{tabular}{lllr}
\toprule
\textbf{Language} & \textbf{Type} & \textbf{Source} & \textbf{Pairs} \\
\midrule
Estonian & natural & Language Help & 46K \\
Estonian & natural & EstNER & 17K \\
Estonian & natural & EstCOPA & 800 \\
Estonian & natural & EIC & 200 \\
Estonian & synthetic & EuroLLM & 175 \\
\midrule
English & natural & Flan V2 & 58K \\
English & natural & Winogrande-m & 2.5K \\
English & synthetic & EuroLLM & 7.2K \\
English & synthetic & Magpie Pro & 2.3K \\
English & synthetic & Tulu V3 & 556K \\
\midrule
English and Estonian & natural translation &  NLLB and MALA & 20K \\
English and Estonian & synthetic translation & Magpie Pro      & 36K \\
English and Estonian & synthetic translation & Cosmopedia   & 7K \\
English and Estonian & synthetic translation & DCLM         & 13K \\
\midrule
\textbf{Total} & & & \textbf{766K} \\
\bottomrule
\end{tabular}
\caption{Instruction tuning mixture components and the number of instruction-response pairs in each component.}
\label{tab:sft-synthetic}
\end{table*}

\subsection{Warm-up phase}
\label{app:warmup}

Before the main continued pre-training stage, we perform a short warm-up phase designed to soften the distribution shift before exposing the model to the full Estonian-enriched mixture. The warm-up follows the same overall design principles as the main mixture but uses partially different sources selected with a stronger emphasis on data quality.

For Estonian, we used curated ENC 2023 subcorpora consisting of the balanced, reference, Wikipedia, and academic subsets. English replay is again represented by Cosmopedia \citep{benallal2024smollmcorpus}, using a disjoint document subset from the main mixture. For mathematics, we use NuminaMath-CoT \citep{numina_math_datasets}, while code data is drawn from Python-Edu \citep{benallal2024smollmcorpus}. Instruction data consists of the top 300k shorter conversations from Magpie-Ultra \citep{xu2025magpie}, synthesized using Llama 3.1 405B. Finally, we include Estonian--English parallel data from NLLB \citep{nllbteam2022languageleftbehindscaling}, filtered using COMET \citep{rei-etal-2022-cometkiwi}.
The composition of the warm-up mixture is summarized in Table \ref{tab:warmup-mixture}.

\subsection{Instruction Tuning Data Details}
\label{app:sft-data}
The supervised fine-tuning mixture combined Estonian and English instruction-following data.

\paragraph{Language help.}
Language help is a corpus of approximately 46K Estonian question-answer pairs collected from a public language support service hosted by the Institute of the Estonian Language (EKI)\footnote{\url{https://eki.ee/keeleinfo/keelenou/annamekeelenou/}} where professional linguists respond to user questions about Estonian language usage. The dataset  provides naturally occurring expert-authored  instruction-following interactions in Estonian and was included as a source of naturalistic supervision for instruction tuning. Minimal preprocessing was applied, including removal of personal names identifying both the question authors and responding linguists.
The data was obtained directly from the EKI with permission for research use, including releasing the models under their respective licenses. However, the dataset itself is not publicly available. 
%To preserve anonymization during review, we do not disclose the data provider in the current submission.

\paragraph{Parallel and benchmark-derived Estonian data.}
In addition to Language help, the Estonian instruction-following mixture included parallel translation data from the NLLB \citep{nllbteam2022languageleftbehindscaling} and Bilingual MaLA \citep{ji2024emma500enhancingmassivelymultilingual} datasets, filtered for quality and length. Existing Estonian NLP datasets---EstCOPA \citep{kuulmets_estcopa_2022}, EstNER \citep{sirts-2023-estonian}, and an Estonian grammar correction dataset\footnote{\url{https://github.com/tlu-dt-nlp/EstGEC-L2-Corpus}}---were additionally converted into instruction-following format and incorporated into the supervised fine-tuning mixture.

\paragraph{Synthetic Estonian instruction data}
To increase the volume of Estonian instruction-following data, we generated synthetic examples using Gemma 3 27B \citep{gemmateam2025gemma3technicalreport}. The synthetic data included translated Magpie-Llama-3.1-Pro prompts \citep{xu2025magpie} used to elicit Estonian responses from the same model, Cosmopedia-style prompt--response pairs, and document-level translations of DCLM documents \citep{li2024datacomp}. Exact dataset sizes and composition statistics are summarized in Table~\ref{tab:sft-synthetic}.

\paragraph{English instruction-following data.}
To preserve broader instruction-following and English-language capabilities during post-training, we incorporated the Tülu-3 supervised fine-tuning mixture \citep{lambert2025tulu} together with synthetic instruction datasets from the EuroLLM project \citep{martins2025eurollm9btechnicalreport}. These datasets cover a broad range of instruction-following tasks, including reasoning, mathematics, coding, and safety-related behaviors.

\section{Training details}
\subsection{Pre-training data preprocessing}
\label{app:cpt-preproc}

To ensure consistent and reproducible production of data mixes, we employed Luigi,\footnote{\url{https://github.com/spotify/luigi}} a lightweight workflow manager with HPC support that provides task dependency tracking and pipeline visibility without significant overhead. The pipeline is fully configured via config files and uses seeded sampling, ensuring that re-runs yield identical outputs. It proceeds in the following stages: documents are sampled from each source by shuffling and slicing at predetermined indices, with light filtering applied; sequences are then tokenized and chunked by token count; finally, a \textbf{best-fit bin-packing} algorithm fills fixed-length buckets to maximize utilization and minimize padding, with individual sequences delimited by an end-of-text token. Unlike the standard concatenate-then-split approach, this preserves document integrity by eliminating unnecessary truncation, which has been shown to improve downstream performance and reduce hallucination at negligible cost to training efficiency \citep{ding2024fewer}. The resulting Arrow dataset, consisting of tensors of uniform maximum-length token sequences is consumed directly by HuggingFace \texttt{datasets}, which serves as the data-loading backend for Accelerate during training.

\subsection{Continued pre-training details}
\label{app:cpt-hyperparams}
The models were trained using HuggingFace \texttt{transformers} \citep{wolf-etal-2020-transformers} and Accelerate \citep{accelerate} using FSDP \citep{10.14778/3611540.3611569}. Training was conducted on the LUMI Supercomputer with 16--32 nodes (8 GPU compute units per node -- 4 AMD MI250x), keeping batch size constant with gradient accumulation.
Hyperparameters for continued pre-training are listed in Table~\ref{tab:cpt-hp}. The learning rate was chosen to be 5e-5 from \{2.5e-5, 5e-5, 1e-4\} based on the held-out validation loss after 1 epoch of \texttt{Llama-3.1-8B} CPT.

The continued pre-training for our \texttt{Llama-3.1-8B}-based model took 10592 GPU-hours (2656 for warmup and 7936 for epoch 1) on 16 nodes, and our \texttt{Apertus-8B}-based model took 15744 GPU-hours (3968 for warmup and 11776 for epoch 1) on 32 nodes. The total computational budget for all experiments was roughly 100,000 GPU-hours.

\begin{table}[!htp]\centering
\small
\begin{tabular}{lrr}\toprule
\textbf{Hyperparameter} &\textbf{Llama-3.1-8B} &\textbf{Apertus-8B} \\\midrule
Learning rate &5e-5 & \\
Optimizer &AdamW & \\
Adam $\epsilon$ &1e-8 & \\
Adam $\beta_1$  &0.9 & \\
Adam $\beta_2$ &0.95 & \\
Sequence length &4096 & \\
Weight decay &0.1 & \\
Scheduler &WSD & \\
FSDP Strategy &\texttt{SHARD\_GRAD\_OP} & \\
Flash attention &\textit{yes} & \\
Grad. checkpoint &\textit{yes} & \\
Precision &bfloat16 & \\
GPUs &128 &256 \\
Nodes &16 &32 \\
Gradient acc &8 &2 \\
Local batch size &1 &2 \\
Global batch size &1024 & \\
- tokens &4194304 & \\
Warm-up steps &2868 &2963 \\
Decay steps &2925 &2960 \\
Total steps &11425 &11625 \\
Epochs &1 & \\
Total training tokens &47.9B &48.8B \\
\bottomrule
\end{tabular}
\caption{Continued pre-training (CPT) hyperparameters. Apertus-8B CPT uses the same setting as Llama-3.1-8B if not stated otherwise.}\label{tab:cpt-hp}

\end{table}

\subsection{Post-training details}
\label{app:posttrain-hyperparams}
Instruction-following fine-tuning was performed via supervised fine-tuning (SFT) using HuggingFace Accelerate distributed across 4 nodes (32 GPUs), employing Flash Attention 2, gradient checkpointing, and bfloat16 mixed precision. The model was trained for one epoch on the SFT mix described in Section~\ref{sec:sft-data} with a learning rate of 2.0e-5, a per-device batch size of 1, 32 gradient accumulation steps, and a maximum sequence length of 4096 tokens. No ablations were performed over the number of training epochs.

Preference optimization was performed using HuggingFace Accelerate distributed across 4 nodes (32 GPUs), employing the same infrastructure and mixed precision settings as the SFT stage. The learning rate was set to 5.0e-7; all other hyperparameters matched the SFT stage. The choice of dataset and method follows \citet{zosa2025continued}.
SFT took 240 GPU-hours and DPO took 26 GPU-hours on the LUMI Supercomputer with 32 GPU compute units (AMD MI250x).

\section{Preliminary CPT experiments}
\label{sec:appenidx-prelim-cpt}

\subsection{Training Epochs}
\label{sec:cpt-epochs}
 \begin{table}[!h]\centering
 \small
 \begin{tabular}{lccccc}\toprule
 \multirow{2}{*}{\textbf{Epochs}} &\multicolumn{2}{c}{\textbf{Disc.} (\textit{Avg acc \%})} &\multicolumn{2}{c}{\textbf{MT} (\textit{BLEU})} \\\cmidrule(lr){2-3}\cmidrule(lr){4-5}
 &\textbf{ET} &\textbf{EN} &\textbf{EN-ET} &\textbf{ET-EN} \\\midrule
 No training &54.0 &76.9 &13.5 &33.7 \\
 Epoch 1 &69.5 &75.4 &28.1 &36.8 \\
 Epoch 2 &70.2 &75.5 &27.0 &36.4 \\
 Epoch 3 &70.4 &74.4 & 28.2 & 36.4\\
 \bottomrule
 \end{tabular}
 \caption{Repeating Estonian data: average accuracy of discriminative benchmarks and FLORES-200 BLEU scores (full results in Table~\ref{tab:cpt-epochs-full}). }\label{tab:base-epochs}
 \end{table}
 
Table~\ref{tab:base-epochs} shows that repeating Estonian training data beyond one epoch provides no substantial additional benefits.

\begin{table}[th]\centering
%\resizebox{\textwidth}{!}{ % use this if the table is too large
\small
\begin{tabular}{llcccc}\toprule
\textbf{} &\textbf{} &\multicolumn{4}{c}{\textbf{Epoch}}  \\\cmidrule{3-6}
\multicolumn{2}{c}{\textbf{Benchmark}} &\textbf{0} &\textbf{1} &\textbf{2} &\textbf{3} \\\midrule
\multicolumn{6}{l}{\cellcolor[HTML]{f3f3f3}\textbf{Discriminative} (\textit{acc})} \\
\multirow{8}{*}{\textbf{ET}} &\textbf{Exam} &44.7 &57.0 &57.5 &57.2 \\
&\textbf{Belebele} &67.0 &77.2 &75.7 &77.6 \\
&\textbf{GlobalPIQA} &53.0 &78.0 &79.0 &81.0 \\
&\textbf{Grammar} &65.8 &87.5 &89.7 &85.0 \\
&\textbf{Declension} &58.7 &61.9 &62.5 &65.9 \\
&\textbf{Trivia} &30.0 &44.9 &46.9 &46.3 \\
&\textbf{WinoGrande} &59.6 &74.0 &72.7 &74.1 \\
&\textbf{XCOPA} &53.2 &75.2 &77.2 &75.8 \\\midrule
\multirow{3}{*}{\textbf{EN}} &\textbf{Belebele} &87.3 &87.0 &87.1 &86.1 \\
&\textbf{Winogrande} &78.5 &76.6 &76.0 &75.4 \\
&\textbf{MMLU-Redux} &64.9 &62.7 &63.4 &61.8 \\\midrule
\multicolumn{6}{l}{\cellcolor[HTML]{f3f3f3}\textbf{Generative} (\textit{BLEU})} \\

\multirow{2}{*}{\textbf{MT}} &\textbf{EN-ET} &13.5 &28.1 &27.0 &28.2 \\
&\textbf{ET-EN} &33.7 &36.8 &36.4 &36.4 \\
\bottomrule
\end{tabular}
\caption{Benchmark results depending on how many epochs the Estonian training data was repeated (\texttt{Llama-3.1-8B}). \textit{Epoch=0} is the base model before CPT. 
%Full results in Appendix~\ref{sec:cpt-epochs} Table~\ref{tab:cpt-epochs-full}.
}
\label{tab:cpt-epochs-full}
\end{table}

\subsection{Choice of Estonian dataset}
\label{sec:dataset-ablation}
\begin{table}[!htp]\centering
\small
\begin{tabular}{lrrrr}\toprule
\textbf{Training DS} &$|D|$ &\textbf{Avg Acc.} \textit{\%} &\textbf{Avg Rank} \\\midrule
\textit{no training} &0 &54.1 &3.1 \\\midrule
ENC &8.6B &\textbf{62.2} &\textbf{1.7} \\
FW2 &15.9B &61.3 &2.9 \\
ET FW mix &19.1B &59.6 &2.1 \\
\bottomrule
\end{tabular}
\caption{Estonian discriminative benchmark scores after continued pretraining with Estonian corpora (monolingual). $|D|$ is the dataset size in Llama 3.1 tokens. Full results in Table~\ref{tab:full-data-ablation}.
}\label{tab:prelim-mono-data}

\end{table}
We experimented with training on monolingual Estonian data mixtures and found that training with ENC yields the highest results on discriminative tasks compared to Fineweb-2 \citep{penedo2025fineweb2} and our own web data mix (ET FW mix), even though those corpora are significantly larger (see Table~\ref{tab:prelim-mono-data}). This indicates that ENC has higher-quality content for LM training, as confirmed by manual inspection of FW2, which reveals a high number of low-quality machine-translated websites. With this in mind, we chose to use ENC as our Estonian corpus in subsequent experiments.

For our own web data mix (ET FW mix), we apply an adapted FineWeb filtering and deduplication pipeline \citep{penedo2024fineweb} to ENC and Estonian web data from CulturaX \citep{nguyen-etal-2024-culturax}, MADLAD-400 \citep{kudugunta2023madlad}, MaLA \citep{ji2024emma500enhancingmassivelymultilingual}, HPLT V2 \citep{burchell-etal-2025-expanded}, and selected snapshots of Community Oscar \citep{brack-etal-2024-community}.

The full results detailing the Estonian discriminative benchmark results for models with different Estonian datasets are in Table~\ref{tab:full-data-ablation}.

\begin{table*}[!htp]\centering
\small
\addtolength{\tabcolsep}{-3pt}
\begin{tabular}{lcccccccccccc}\toprule
\textbf{Dataset} &$|D|$ &\textbf{Exam} &\textbf{Belebele} &\textbf{Grammar} &\textbf{Declension} &\textbf{Trivia} &\textbf{XCOPA} &\textbf{WinoGrande} & &\textbf{Avg Acc} &\textbf{Avg Rank} \\\midrule
\textit{no training} &0 &44.7 &67.0 &65.8 &58.7 &30.0 &53.2 &59.6 & &54.1 &3.1 \\\midrule
ENC &8.6B &54.6 &67.0 &72.1 &48.2 &45.6 &75.4 &72.8 & &62.2 &1.7 \\
FW2 &15.9B &53.4 &65.6 &71.6 &48.0 &41.9 &74.4 &74.2 & &61.3 &2.9 \\
ET FW mix &19.1B &55.1 &67.4 &59.0 &42.7 &43.4 &75.0 &74.5 & &59.6 &2.1 \\
\bottomrule
\end{tabular}
\caption{Estonian discriminative benchmark accuracies (\%) after continued pretraining with Estonian corpora (monolingual).}\label{tab:full-data-ablation}

\end{table*}

\subsection{Base model merging}
\label{sec:slerp-ablation}
\begin{figure*}[!htp]
    \centering
    \includegraphics[width=\linewidth]{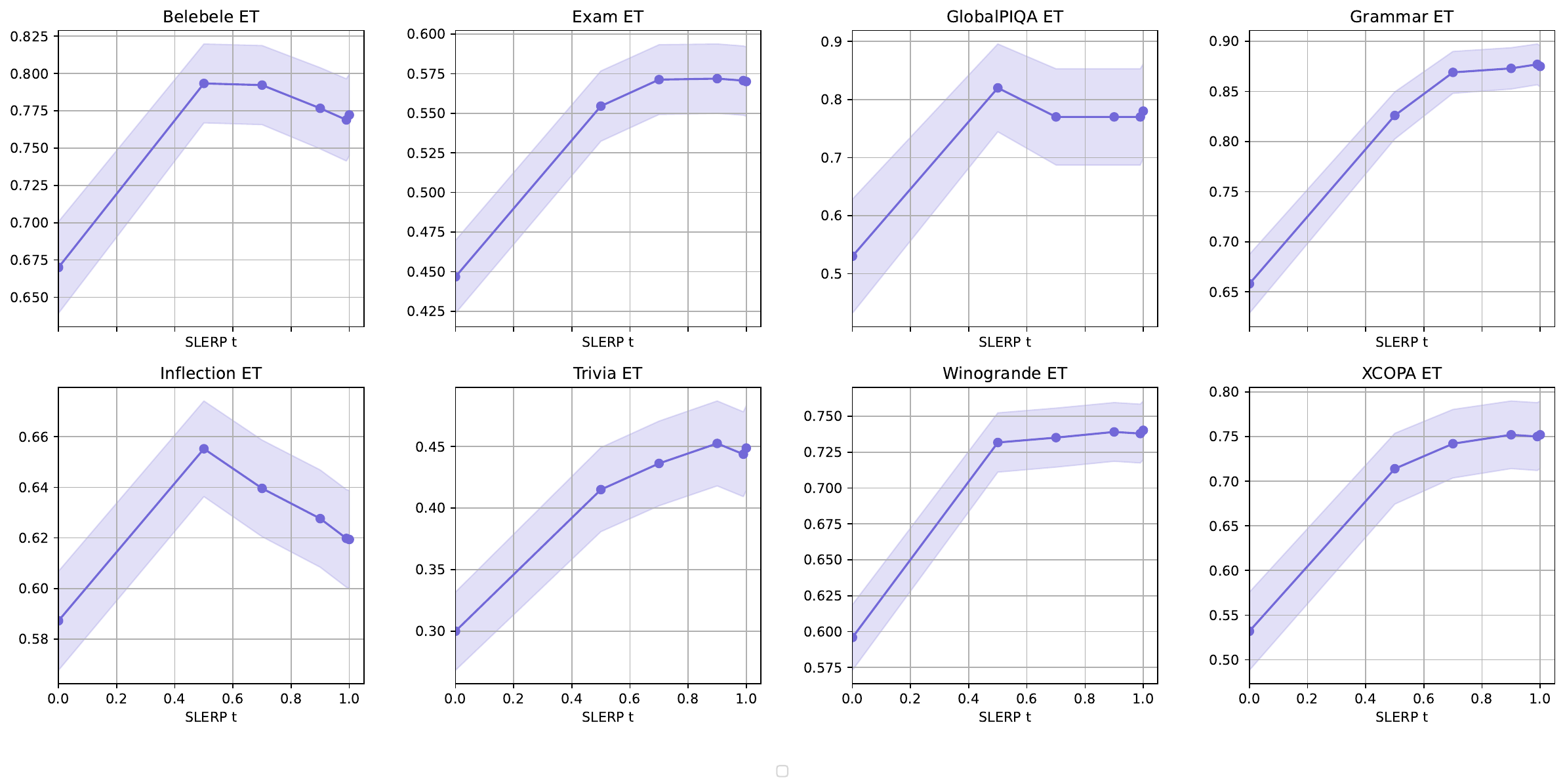}
    \includegraphics[width=0.8\linewidth]{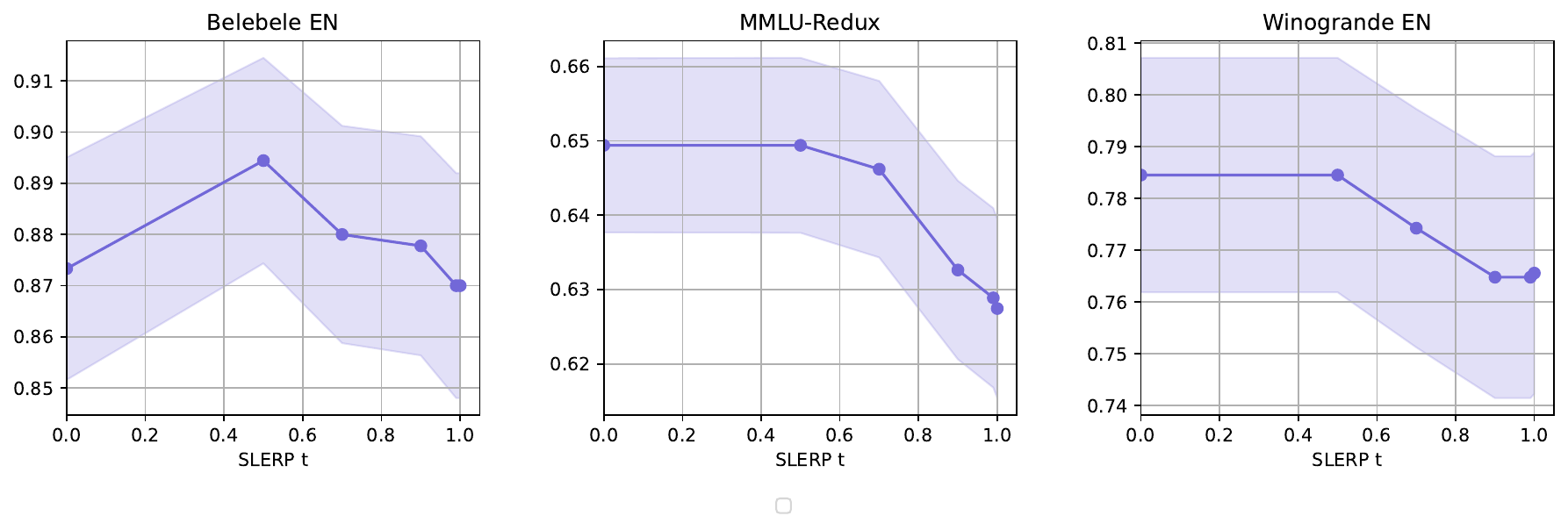}
    \includegraphics[width=0.55\linewidth]{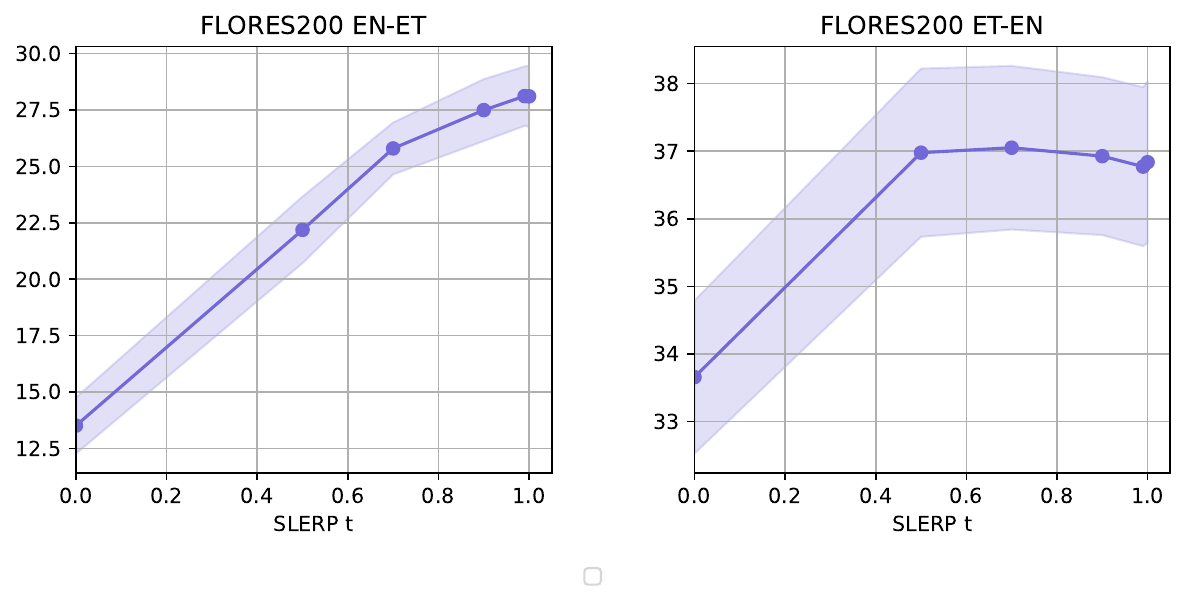}
    \caption{Results of base model SLERP merging, where $t=0$ is \texttt{Llama-3.1-8B} and $t=1$ is \texttt{Llama-3.1-EstLLM-8B}. The colored area represents the 95\% confidence interval calculated from the standard error reported by \texttt{LM evaluation harness} \citep{lintang_sutawika_2024_12608602}.}
    \label{fig:slerp}
\end{figure*}

We experimented with merging the \texttt{Llama-3.1-8B} model with \texttt{Llama-3.1-EstLLM-8B} using SLERP from \texttt{mergekit} \citep{goddard-etal-2024-arcees} (see Figure~\ref{fig:slerp}). We observed improvements on some discriminative tasks; however, our only generative task---machine translation on FLORES200---deteriorated, so we decided not to pursue this approach further.

\section{Pairwise Human Evaluation}
\label{app:human_eval}

\subsection{Dataset construction}
\label{app:dataset_construction}
We construct a new curated multi-turn Estonian benchmark for targeted pairwise human evaluation. The benchmark consists of 80 two-turn conversations translated from LMSYS-Chat \citep{zheng2024lmsys}, reusing 30 translations from SMUGRI-MT-Bench \citep{purason-etal-2025-llms} with minor modifications. A second user turn was manually added to each conversation.

The benchmark covers four categories:
\begin{itemize}[nosep]
    \item planning and scheduling,
    \item specific-format writing,
    \item mathematics,
    \item logical reasoning.
\end{itemize}

The benchmark was designed to evaluate both conversational quality and multi-turn instruction following in Estonian.

\subsection{Evaluation pair construction}
\label{app:eval_pair_construction}
The human evaluation was designed around a set of targeted research questions comparing specific adaptation decisions. Seven instruction-tuned model variants were included in the evaluation. For each of the 80 benchmark conversations, we generated 5 different responses from the relevant models using temperature 0.4. Since each conversation consisted of two turns, generations were conditioned on the full conversation history.

The generated responses were organized into pairwise comparisons corresponding to six targeted research questions, such that in each comparison only a single adaptation decision differed between the evaluated models (e.g., replay, repeated epochs, chat vector merging, or base model choice). Each research question contained 400 response pairs (80 conversations × 5 sampled generations), yielding 2400 response pairs in total across all six comparison settings.

The response pairs were divided into 100 annotation batches of 24 comparisons each. Each batch contained at most one comparison derived from a given benchmark conversation to avoid repeated exposure to the same prompt content.

\subsection{Annotation procedure}

Ninety-nine undergraduate computer science students signed up for the human evaluation study and seventy-four students completed the task. The study took place in an artificial intelligence course taught in Estonian, and 95\% of the participants self-reported Estonian as their first language. 
Participation was voluntary, and students received additional course credit for completing the task.
Students were informed that their contributions would be used to evaluate the Estonian capabilities of several LLMs.
 
Each student annotated one batch of 24 response pairs constructed as described in previous section~\ref{app:eval_pair_construction}. Thus, in total, 74 batches comprising 1776 response pairs were annotated. Annotations were based on the full two-turn conversation. The model names were anonymized, and the response order was randomized independently for each comparison. 

\begin{figure*}[h!]
    \includegraphics[width=0.92\textwidth]{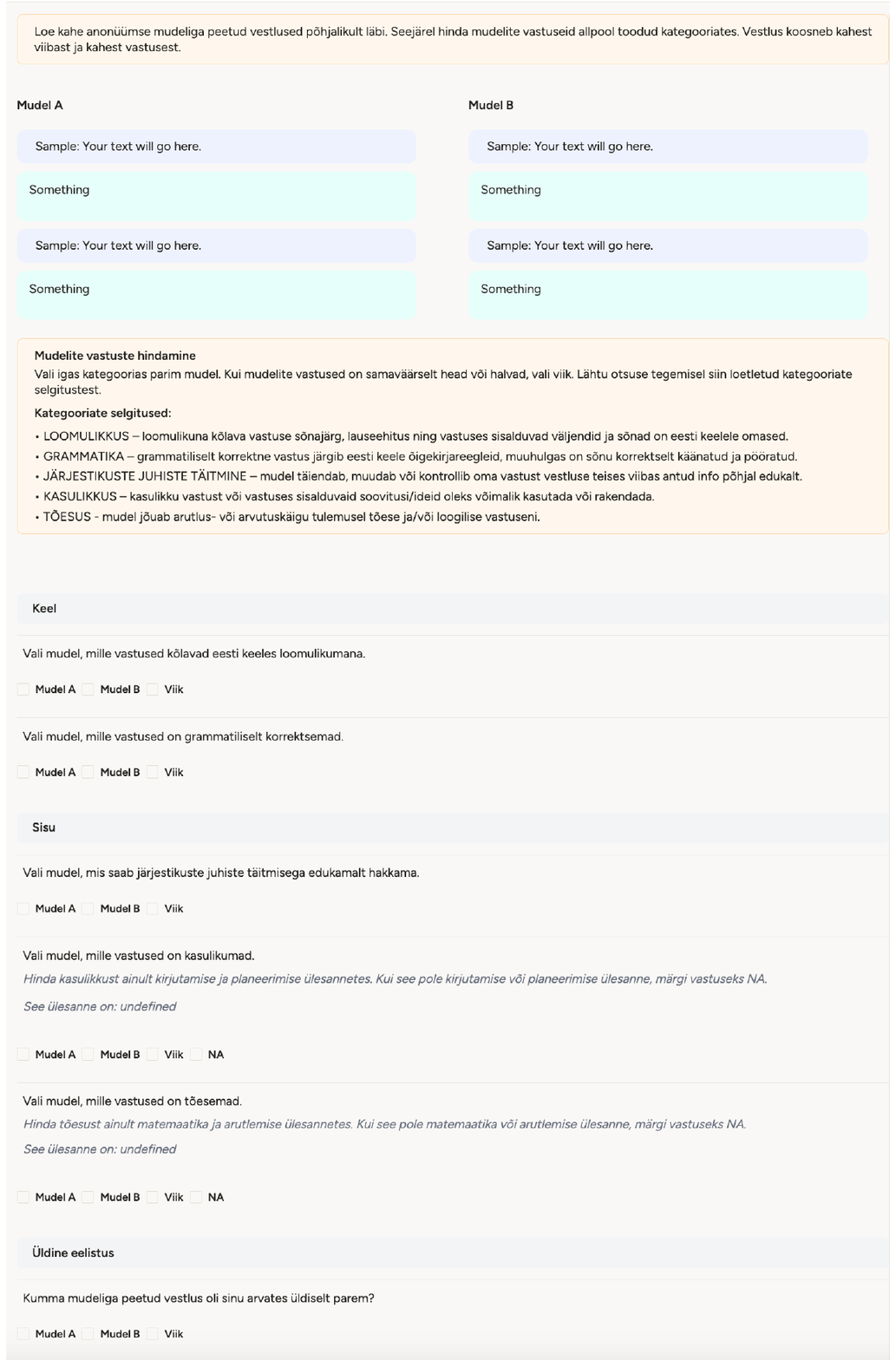}
    \caption{Annotation interface used for human evaluation. The interface provides task instructions and definitions of the evaluation categories (shown in yellow), side-by-side model responses (shown in teal) to two consecutive prompts (shown in blue), and checkboxes for indicating preferences in each category. Original interface text is shown in Estonian. The English translation of the instructions is as follows: “\textit{Read the conversations held with two anonymous models thoroughly. Then evaluate the models’ responses in the categories provided below. A conversation consists of two prompts and two responses. Choose the better model in each category. If the models’ responses are equally good or equally bad, choose a tie. Base your decision on the definitions of the categories listed here.}”}
    \label{fig:evaluation-interface}
\end{figure*}

Annotations were collected using Label Studio \citep{LabelStudio}. The annotation interface (Figure \ref{fig:evaluation-interface}) presented side-by-side model responses together with instructions, definitions of the evaluation criteria, and checkboxes for recording preferences. All information on the interface was provided in Estonian.

Students were instructed to read both conversations fully and evaluate which model performed better in terms of fluency, grammatical correctness, and sequential instruction following. For writing and planning tasks, students additionally evaluated usefulness, while for math and reasoning tasks, they evaluated correctness. Students also recorded their overall preference between the two conversations. Responses were required for all relevant evaluation categories and could not be skipped. If both conversations were considered equally good or equally poor in a given category, students were instructed to select a tie.
Annotation time was recorded for all participants. The median completion time was 2 hours and 24 minutes. 

\subsection{Additional results}
\begin{figure}[h!]
    \centering
    \includegraphics[width=\linewidth]{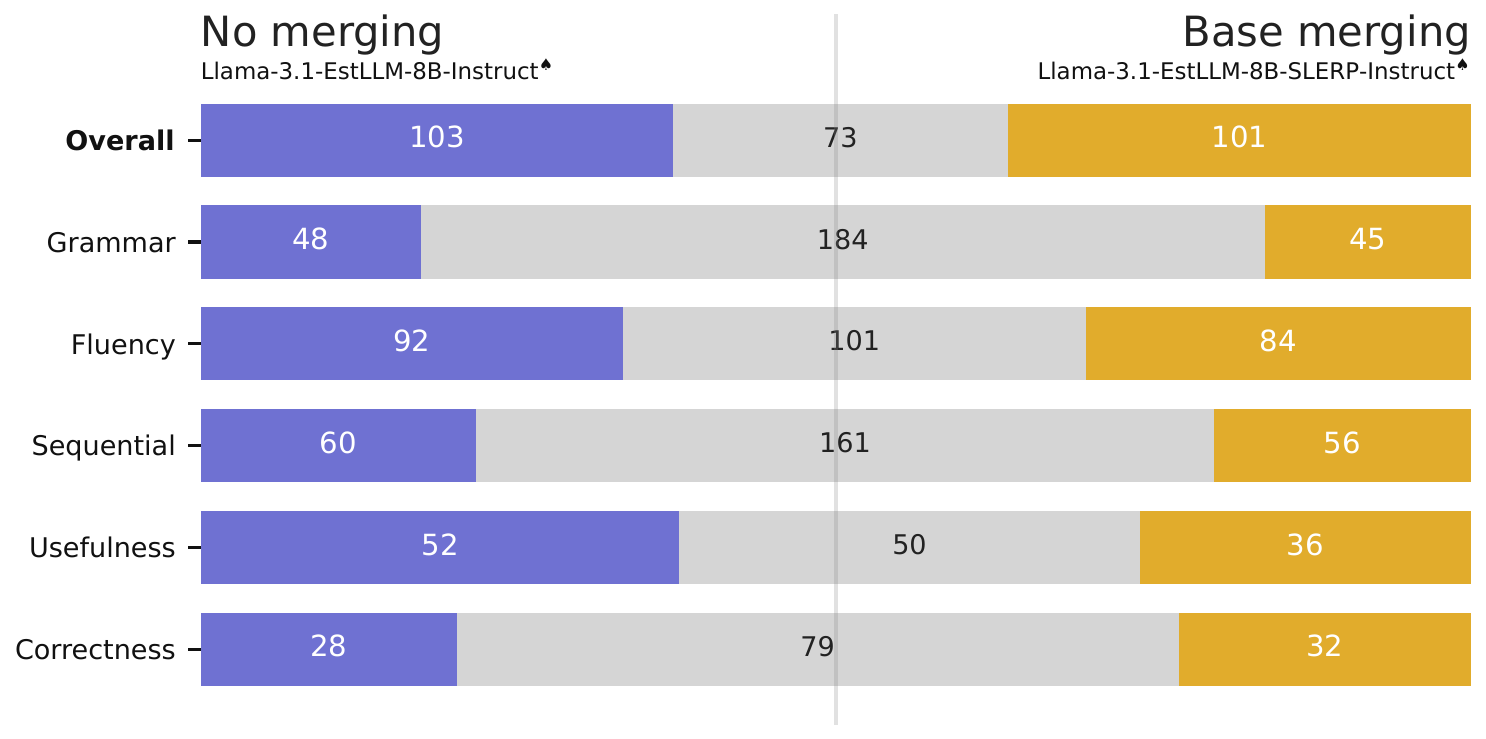}
    \caption{Pairwise human evaluation results: \textit{Does base model merging improve the results?} \textsuperscript{$\spadesuit$} denotes chat-vector merging.}
    \label{fig:human-eval-add-cptmerge}
\end{figure}

\begin{figure}[h!]
    \centering
    \includegraphics[width=\linewidth]{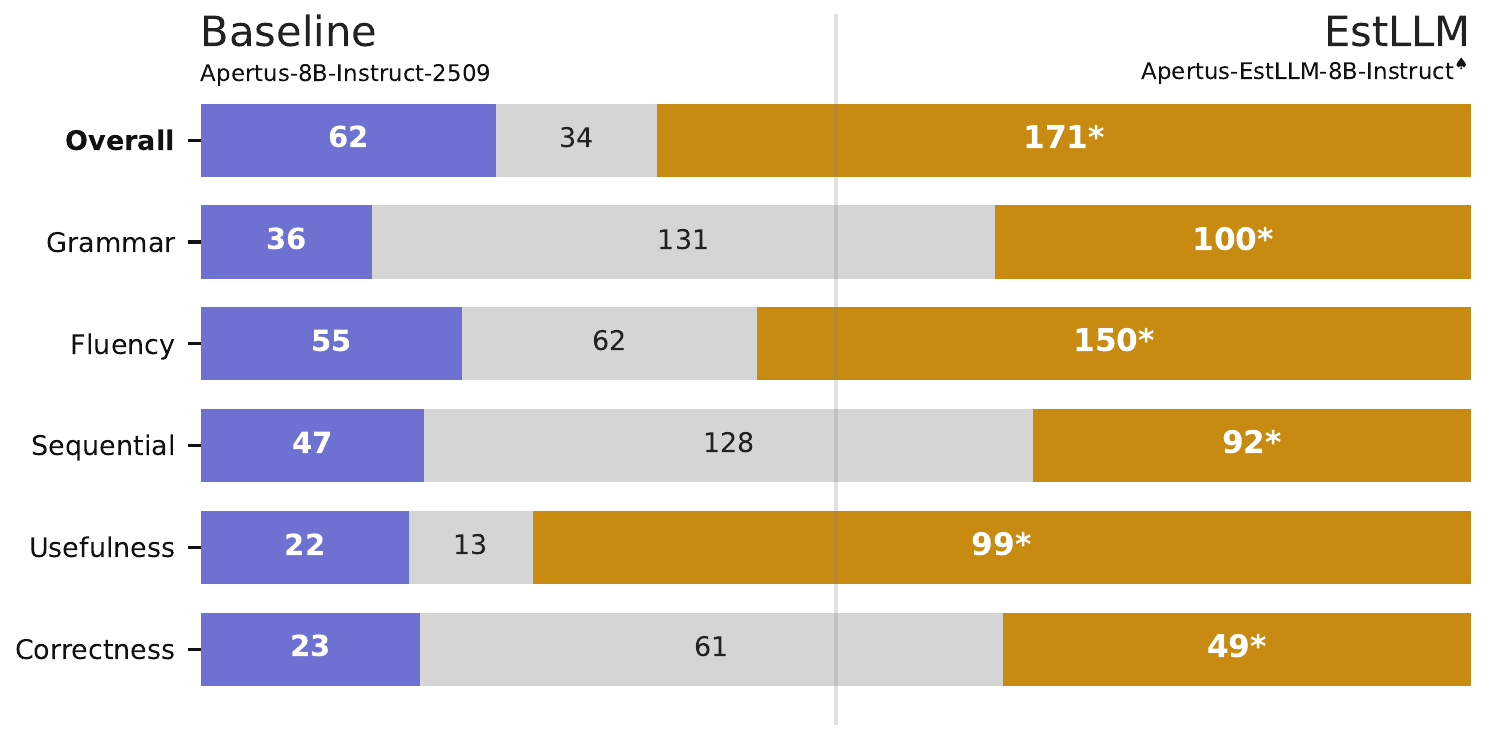}
    \caption{Pairwise human evaluation results: \textit{Does our pipeline improve \texttt{Apertus-8B} compared to \texttt{Apertus-8B-Instruct}?} Significant results indicated with * ($\alpha=0.05$). \textsuperscript{$\spadesuit$} denotes chat-vector merging.}
    \label{fig:human-eval-apertus}
\end{figure}

We present additional results for pairwise human evaluation in Figures \ref{fig:human-eval-add-cptmerge} and \ref{fig:human-eval-apertus}.

\paragraph{\faQuestionCircle{} Does base model merging improve the results?} Human evaluation shows no evidence that merging \texttt{Llama-3.1-EstLLM-8B} base model with \texttt{Llama-3.1-8B} using SLERP provides any advantage across any of the categories (see Figure~\ref{fig:human-eval-add-cptmerge}).

\paragraph{\faQuestionCircle{} Does our pipeline improve \texttt{Apertus-8B} compared to \texttt{Apertus-8B-Instruct}?} Human evaluation shows that our pipeline significantly improves on the baseline Apertus model across all categories (see Figure~\ref{fig:human-eval-apertus}).

\section{Arena-style Human Evaluation}
\label{app:arena-results}

The Arena-style evaluation platform was implemented by adapting the publicly available LMSYS Chatbot Arena codebase \citep{chiang2024chatbot} and translating the interface into Estonian. The platform provided access to both proprietary and open-source Estonian-capable models and was publicly available for approximately nine months at the time of the reported snapshot. Instruction-tuned EstLLM checkpoints had been available on the platform for approximately 2--4 months, depending on the checkpoint. In total, the platform included more than 50 evaluated models and accumulated approximately 15K pairwise preference votes. 

Model scores were estimated from pairwise preferences using the Bradley-Terry model. The 95\% confidence intervals were calculated by fitting the model to 100 bootstrap samples drawn with replacement from the original evaluation data. Each model's rank was calculated as one plus the number of models whose lower confidence bound exceeded that model's upper confidence bound.
A snapshot of the top-20 ranked open models is shown in Table~\ref{tab:arena_eval}.

\section{Model Release and Licensing}
\label{app:model-release}

We release the resulting EstLLM models under the licenses of the respective base models. Llama 3.1 based EstLLM models are released under the original \textit{Llama 3.1 license}, while Apertus-based EstLLM models follow the original \textit{Apache-2.0} license of Apertus. This is consistent with the licenses, permissions, and intended use of the original models and the training datasets used in this work.

\section{Evaluation details}
\label{app:eval-details}
The number of few-shot examples for base model evaluation is specified in Table~\ref{tab:eval-nfs-base}.

\begin{table}[!htp]\centering\small

\begin{tabular}{lrr}\toprule
\textbf{Benchmark} &\textbf{Num fewshot} \\\midrule
Belebele (EN/ET) &5 \\
Exam &5 \\
FLORES &5 \\
GlobalPIQA &0 \\
Grammar &5 \\
Declension &5 \\
Trivia &5 \\
Winogrande (EN/ET) &3 \\
XCOPA &5 \\
MMLU-Redux &0 \\
\bottomrule
\end{tabular}
\caption{Number of few-shot examples in base model evaluation. FLORES200 is evaluated with deterministic decoding, while the rest are not evaluated generatively.}\label{tab:eval-nfs-base}

\end{table}

\begin{table*}[!tp]\centering
%\resizebox{\textwidth}{!}{ % use this if the table is too large
\begin{tabular}{lllrrrr}
\toprule
\bf No & \bf Rank & \bf Model & \bf Model size & \bf Score & \bf 95\% CI & \bf Votes \\
\midrule
1 & 1 & DeepSeek-V3 (0324) & 685B & 1457 & $-$28/$+$41 & 964 \\
2 & 1 & Kimi-K2-Instruct & 1T & 1442 & $-$40/$+$36 & 729 \\
3 & 1 & Llama 4 Maverick & 402B & 1432 & $-$31/$+$38 & 749 \\
4 & 1 & Gemma-3-27B-it & 27B & 1413 & $-$38/$+$35 & 750 \\
5 & 1 & Llama-EstLLM-8B-Instruct\textsuperscript{$\spadesuit$} & 8B & 1384 & $-$47/$+$53 & 110 \\
6 & 2 & Llama-EstLLM-8B-Instruct & 8B & 1376 & $-$40/$+$39 & 541 \\
7 & 2 & Llama 4 Scout & 109B & 1374 & $-$37/$+$32 & 753 \\
8 & 4 & Meta-Llama-3.1-405B-Instruct & 405B & 1358 & $-$35/$+$38 & 624 \\
9 & 4 & Qwen3-235B-A22B & 235B & 1347 & $-$40/$+$36 & 661 \\
10 & 4 & Apertus-EstLLM-8B-Instruct & 8B & 1321 & $-$45/$+$57 & 164 \\
11 & 4 & Gemma-3-12B-it & 12B & 1317 & $-$65/$+$59 & 155 \\
12 & 5 & Llama-3.3-70B-Instruct & 70B & 1329 & $-$33/$+$37 & 840 \\
13 & 5 & EuroLLM-9B-Instruct & 9B & 1328 & $-$32/$+$43 & 437 \\
14 & 8 & Apertus-8B-Instruct-2509 & 8B & 1285 & $-$46/$+$42 & 283 \\
15 & 8 & Meta-Llama-3.1-70B-Instruct & 70B & 1281 & $-$46/$+$52 & 230 \\
16 & 9 & Mistral Small 3.1 (2503) & 24B & 1270 & $-$27/$+$37 & 742 \\
17 & 12 & Qwen3-32B & 32B & 1251 & $-$33/$+$38 & 586 \\
18 & 12 & Llammas & 7B & 1243 & $-$41/$+$44 & 406 \\
19 & 13 & Qwen2.5-72B-Instruct & 72B & 1228 & $-$36/$+$34 & 499 \\
20 & 13 & Llama-3-70B-Instruct & 70B & 1222 & $-$28/$+$39 & 598 \\
\bottomrule
\end{tabular}
\caption{Top-20 ranked open models leaderboard from the Chatbot Arena style evaluation page in Estonian. \textsuperscript{$\spadesuit$} denotes chat-vector merging.}
\label{tab:arena_eval}
\end{table*}

\end{document}